\documentclass[]{gtech}
\usepackage{amssymb}
\usepackage{multirow}
\usepackage{bigdelim}
\usepackage{todonotes}
\usepackage{longtable}
\usepackage{tabularray}
\usepackage{wrapfig}
\usepackage[most]{tcolorbox} 
\usepackage{xcolor}
\usepackage{url}
\usepackage{hyperref} 

\usepackage{fontawesome5} 
\usepackage{hyperref}     
\usepackage{pifont}
\usepackage{float}
\usepackage{tikz}
\usepackage{marvosym}
\usepackage{bbding}
\usepackage[table]{xcolor}
\usepackage{tabularx}
\usepackage{booktabs}
\usepackage{tcolorbox}
\usepackage{adjustbox}
\usepackage{makecell}
\usepackage{multirow}
\usepackage{pifont}
\usepackage{arydshln}
\usepackage{graphicx}     
\usepackage{multirow}     
\usepackage[table]{xcolor}
\usepackage{hhline}       

\definecolor{darkblue}{RGB}{94,110,186}
\newcommand{\darkblue}[1]{\textcolor{darkblue}{#1}}

\renewcommand{\title}[1]{\newcommand{\titlelist}{{\huge\fontfamily{optimistic}\selectfont #1}}}

\newcommand{\baseline}{\texttt{Ivy-xDetector}}
\newcommand{\benchmark}{\texttt{Ivy-Fake}}

\definecolor{prompt}{HTML}{5f84e4}
\definecolor{img}{HTML}{820100}

\usepackage{arydshln}

\definecolor{CQColor}{rgb}{0.0,0.0,1.0} 

\definecolor{TSColor}{rgb}{0.5,0.0,0.8} 

\definecolor{CQRColor}{rgb}{1.0,0.0,1.0} 

\usepackage{colortbl}
\usepackage{amssymb}
\usepackage{pifont}
\usepackage{booktabs,multirow}
\usepackage{makecell}
\usepackage{tabulary}
\usepackage{fontawesome5}
\usepackage{bbding}
\usepackage{multicol}

\newlength\savewidth

\title{\textcolor[HTML]{E65C00}{Ivy}-Fake: A Uni{f}ied Explainable Framework and Benchmark for Image and Video AIGC Detection}

\author{Changjiang Jiang*} 
\author{Wenhui Dong*\Letter} 
\author{Zhonghao Zhang}
\author{Fengchang Yu}
\author{Wei Peng}
\author{Xinbin Yuan}
\author{Yifei Bi}
\author{Ming Zhao}
\author{Zian Zhou}
\author{Chenyang Si}
\author{Caifeng Shan*\Letter}

\affiliation{Nanjing University\textsuperscript{+}}

\contribution[*]{Equal Contributions}
\contribution[]{\Letter~Corresponding author: wenhui.dong@smail.nju.edu.cn,cfshan@nju.edu.cn}
\contribution[+]{\textit{Complete affiliation can be see in Sec.~\ref{sec:contri}}}

\abstract{\fontsize{11pt}{12pt} \textit{The rapid development of Artificial Intelligence Generated Content (AIGC) techniques has enabled the creation of high-quality synthetic content, but it also raises significant security concerns. Current detection methods face two major limitations: (1) the lack of multidimensional explainable datasets for generated images and videos. Existing open-source datasets (e.g., WildFake, GenVideo) rely on oversimplified binary annotations, which restrict the explainability and trustworthiness of trained detectors. (2) Prior MLLM-based forgery detectors (e.g., FakeVLM) exhibit insufficiently fine-grained interpretability in their step-by-step reasoning, which hinders reliable localization and explanation. To address these challenges, we introduce Ivy-Fake, the first large-scale multimodal benchmark for fake image and video detection. It consists of over 106K richly annotated training samples (images and videos) and 5,000 manually verified evaluation examples, sourced from multiple generative models and real-world datasets through a carefully designed pipeline to ensure both diversity and quality. Furthermore, we propose Ivy-xDetector, a multimodel large language model (MLLM) based on reinforcement fine-tuning (RFT), capable of producing explainable reasoning chains and achieving robust performance across multiple fake image and video detection benchmarks.
}}

\gtechdata[Project Homepage]{%
    \href{https://pi3ai.github.io/IvyFake}{\underline{Link}}\vspace{-1mm}%
}

\gtechdata[Model]{%
    \href{https://huggingface.co/AI-Safeguard/Ivy-Fake}{\url{https://huggingface.co/AI-Safeguard/Ivy-Fake}}%
}

\gtechdata[Dataset]{%
    \href{https://huggingface.co/datasets/AI-Safeguard/Ivy-Fake}{\url{https://huggingface.co/datasets/AI-Safeguard/Ivy-Fake}}%
}

\begin{document}
\maketitle

\section{Introduction}
\label{sec:intro}



\begin{figure*}[ht]
    \centering
    \includegraphics[width=0.95\textwidth]{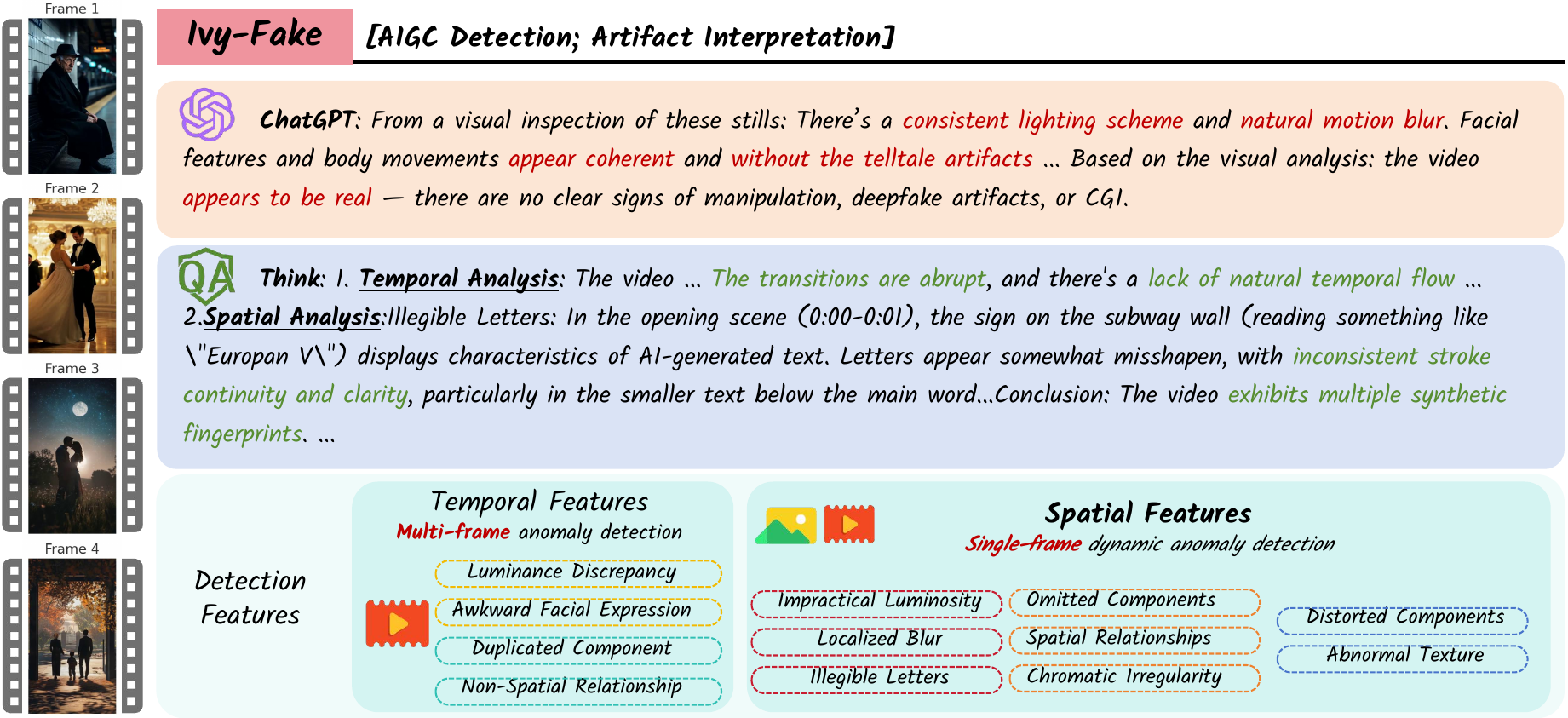}
    \caption{Overview of the \benchmark~framework: By conducting in-depth analysis of temporal and spatial artifacts, the framework enables explainable detection of AI-generated content.}
    \label{fig:figure1-poster} 
\end{figure*}

The rapid development of diffusion-based models has triggered an exponential growth in Artificial Intelligence Generated Content (AIGC), like Sora~\citep{sora_videoworldsimulators2024}, DALL-E~\citep{dalle_3_betker2023improving}, Imagen~\citep{imagegen_saharia2022photorealistic}, and Stable Diffusion~\citep{rombach2022high}, which have redefined state-of-the-art performance in text-to-image synthesis. However, these advances also raise significant security concerns, including the misuse of DeepFake~\citep{Qureshi2024DeepfakeFA}, document tampering~\citep{Qureshi2024DeepfakeFA,qu2023doctamper,qu2024miml,qu2025ostf}, and dataset poisoning~\citep{10.5555/3294996.3295110,ying2026beyond}. The increasing realism of synthetic content blurs the boundary between genuine and fabricated media, posing critical challenges for misinformation control, content provenance verification, and the preservation of public trust.

However, most existing approaches primarily focus on binary authenticity classification~\citep{demamba,wildfake_2025,aigvdet_bai2024ai,aide_yan2025a}, which limits the human interpretability of model predictions~\citep{qu2026textshield,qu2024omni,loki_ye2024loki,fakehr1}. Existing benchmarks also remain inadequate for evaluating explainable AIGC detection. Datasets such as AIGCDetectionBenchmark~\citep{aigcdetectionbenchmark_zhong2023patchcraft} and GenVideo~\citep{demamba} provide only binary labels, while more recent resources like LOKI~\citep{loki_ye2024loki} attempt to incorporate fine-grained anomaly annotations across modalities but are still constrained in scale and diversity. On the other hand, current methods such as AIDE~\citep{aide_yan2025a} and Demamba~\citep{demamba} are limited to a single modality—either image or video—without exploring unified detection across both domains. Similarly, existing datasets exhibit the same fragmentation: FakeBench~\citep{liu2024mmfakebench} emphasizes explainable fake image detection but omits video content, whereas FakeClue~\citep{fakeclue_wen2025spot} provides extensive image-level annotations yet lacks integrated video data. This fragmentation leads to substantial gaps in benchmarking and hinders unified advancement in explainable multimodal AIGC detection.

To address these challenges, we propose \benchmark, a comprehensive dataset designed to evaluate explainable multimodal AIGC detection. \benchmark~offers: 1) Diverse Multimodal Data, a large-scale dataset comprising 61,107 annotated images and 45,272 annotated videos for training, along with 5,000 samples for evaluation. 2) Explainable Annotations, rich annotations that extend beyond binary labels to include detailed reasoning, enabling nuanced evaluation of models' interpretability and explanatory capabilities. 

In addition to providing \benchmark, we introduce \baseline, a model for detecting AI-generated images and videos and explaining the associated artifacts. Unlike other MLLMs, \baseline~excels at spotting generative artifacts, spatial in images and both spatial and temporal in videos. By integrating multiple spatial and temporal feature extractors, it detects image-level artifacts and video-level temporal inconsistencies with superior accuracy. Through the incorporation of explainable annotations and carefully curated domain-diverse data, our approach achieves state-of-the-art performance using only a fraction of the data required by existing detectors. While prior image and video classifiers~\citep{demamba,aide_yan2025a,aigcdetectionbenchmark_zhong2023patchcraft} often rely on millions (M-level) of samples, our method attains superior performance across both image and video AIGC detection tasks with fewer than 200K samples.


Our main contributions are summarized as follows:

\begin{itemize}
\item We introduce \benchmark, the first large-scale dataset for explainable AIGC detection across both images and videos, comprising 106,379 training samples and 5,000 manually verified test instances. Each entry is enriched with fine-grained visual annotations and textual reasoning to support transparent multimodal evaluation and future research.
\item We propose \baseline, a reinforcement learning–based model by Reinforcement Learning with Verifiable Rewards (RLVR) built upon Group Relative Policy Optimization (GRPO), which achieves superior performance on multiple image and video detection benchmarks using fewer than 200K samples, significantly outperforming existing methods that rely on millions of data points.
\item We conduct extensive experiments to validate the effectiveness of both the dataset and the model, demonstrating their substantial impact on improving multimodal large models’ capability in synthetic content detection.
\end{itemize}
\begin{table*}[tp]
    \centering
    \caption{Task examples of \benchmark.}
    \label{tab:task_dimension}
    \resizebox{0.95\textwidth}{!}{
        \begin{tabular}{c|c|c}
        \hline
        \textbf{Features} & \textbf{Sub Dimension} & \textbf{Example} \\
        \hline
        \multirow{16}{*}{\textbf{Spatial}} & Impractical & \cellcolor{gray!5}{\textit{\darkblue{Which visual clue reveals an impossible spatial or lighting configuration in the scene?}}} \\
        ~ & Luminosity & \cellcolor{gray!5}{(A) Light direction opposite source. (B) Face without lighting. (C) Dark background, bright face. (D) Bright background, dark face.} \\
        \hhline{~|-|-}
        ~ & Localized & \cellcolor{gray!5}{\textit{\darkblue{Which clue shows local blur or focus inconsistency?}}} \\
        ~ & Blur & \cellcolor{gray!5}{(A) Clear face but blurred hand. (B) Sharp background, fuzzy hair. (C) Glasses edges blurred, eyes clear. (D) Hand blurred while holding clear object.} \\
        \hhline{~|-|-}
        ~ & Illegible & \cellcolor{gray!5}{\textit{\darkblue{Which part of the document text is tampered or altered?}}} \\
        ~ & Letters & \cellcolor{gray!5}{(A) Text blurred or half missing. (B) Letters inconsistent in size. (C) Words overlap or melt together. (D) Signs contain random symbols.} \\
        \hhline{~|-|-}
        ~ & Omitted & \cellcolor{gray!5}{\textit{\darkblue{Which part of the person looks incomplete or missing?}}} \\
        ~ & Components & \cellcolor{gray!5}{(A) Missing one finger. (B) Two fingers merged. (C) Missing eyebrows. (D) Three pairs of glasses.} \\
        \hhline{~|-|-}
        ~ & Spatial & \cellcolor{gray!5}{\textit{\darkblue{Where does the spatial relation look unnatural or physically impossible?}}} \\
        ~ & Relationships & \cellcolor{gray!5}{(A) Arm passes through body. (B) Legs overlap unnaturally. (C) Head detached from neck. (D) Hand holds object from wrong side.} \\
        \hhline{~|-|-}
        ~ & Distorted & \cellcolor{gray!5}{\textit{\darkblue{Which part appears deformed or stretched in shape?}}} \\
        ~ & Components & \cellcolor{gray!5}{(A) Finger bent at odd angle. (B) Eye stretched sideways. (C) Mouth warped or uneven. (D) Hand larger than face.} \\
        \hhline{~|-|-}
        ~ & Chromatic & \cellcolor{gray!5}{\textit{\darkblue{Which clue shows unnatural or inconsistent colors in the image?}}} \\
        ~ & Irregularity & \cellcolor{gray!5}{(A) Face partly blue under warm light. (B) Green reflection on red shirt. (C) Sky color bleeds into hair. (D) Background color unevenly saturated.} \\
        \hhline{~|-|-}
        ~ & Abnormal & \cellcolor{gray!5}{\textit{\darkblue{What visual area shows abnormal or inconsistent texture?}}} \\
        ~ & Texture & \cellcolor{gray!5}{(A) Skin looks like plastic. (B) Hair pattern repeats unnaturally. (C) Cloth surface too smooth or glossy. (D) Wall texture mixed with skin.} \\
        \hline
        \multirow{8}{*}{\textbf{Temporal}} & Luminance & \cellcolor{gray!5}{\textit{\darkblue{Which moment shows sudden or inconsistent brightness between frames?}}} \\
        ~ & Discrepancy & \cellcolor{gray!5}{(A) Light flickers without reason. (B) Face turns dark while background stays bright. (C) Shadow jumps between frames. (D) Object brightness changes abruptly.} \\
        \hhline{~|-|-}
        ~ & Awkward & \cellcolor{gray!5}{\textit{\darkblue{When does the facial movement look unnatural or discontinuous across frames?}}} \\
        ~ & Facial Expression & \cellcolor{gray!5}{(A) Smile appears and disappears suddenly. (B) Eyes blink in different directions. (C) Mouth freezes mid-motion. (D) Expression changes without emotion cue.} \\
        \hhline{~|-|-}
        ~ & Duplicated & \cellcolor{gray!5}{\textit{\darkblue{Where does an object or body part appear twice across frames?}}} \\
        ~ & Component & \cellcolor{gray!5}{(A) Two faces shown in one frame. (B) Hand shadow moves separately. (C) Extra arm appears then fades. (D) Same object duplicated in next frame.} \\
        \hhline{~|-|-}
        ~ & Non-Spatial & \cellcolor{gray!5}{\textit{\darkblue{When does the action sequence look illogical or out of order?}}} \\
        ~ & Relationships & \cellcolor{gray!5}{(A) Person closes door before touching it. (B) Cup moves without contact. (C) Smile appears after laughter. (D) Object vanishes then reappears suddenly.} \\
        \hline
        \end{tabular}
    }
\end{table*}

\begin{figure*}[htbp]
    \centering
    \includegraphics[width=0.95\textwidth]{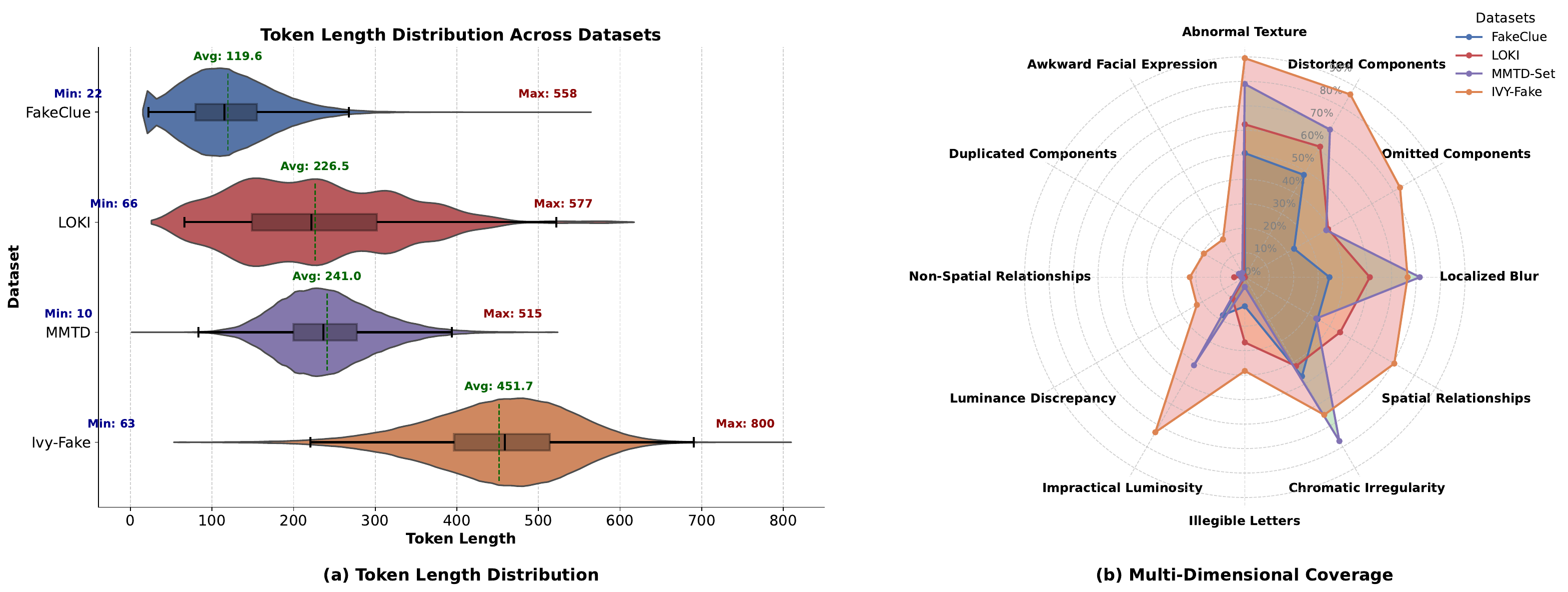}
    \caption{Token Length Distributions and Multi-Dimensional Coverage Across Datasets; Left: Distribution of token lengths across datasets; Right: Coverage of multiple dimensions in explainability datasets, extracted using Qwen3-32B~\citep{qwen3}. The Prompt can be seen in appendix.} 
    \label{fig:figure-data-staic} 
\end{figure*}

\section{Related Work}

\subsection{Methods for Synthetic Content Detection}
Due to growing concerns about the misuse of synthetic data~\citep{deng2024survey,qu2025ostf,li2025artificial}, research on AI-generated content (AIGC) detection has expanded rapidly in recent years~\citep{wang2026forgeryvcr,park2025vidguard,shuai2026detectors,tan2026videoveritas,tan2025veritas,yu2026agentfox,huang2025unishield,Zhou2025AIGIHolmesTE}. Most existing models for AI-generated images and videos formulate the task as binary classification, simply predicting whether the content is "real" or "fake." Representative examples include CNN-based AIGVDet~\citep{aigvdet_bai2024ai}, CNNSpot~\citep{cnnspot_wang2020cnn}, DIRE~\citep{dire_wang2023dire}, Mirror~\citep{liu2026mirror}  and AIDE~\citep{aide_yan2025a}. Meanwhile, several works have explored the application of multimodal large language models (MLLMs) to AIGC detection, including Synartifact~\citep{cao2024synartifact} and Bi-LORA~\citep{keita2025bi}. However, these approaches largely overlook the importance of interpretability in AIGC detection.

Some efforts attempt to introduce interpretability by leveraging spatial annotations~\citep{qu2023doctamper,zhu2025mesorch,qu2024miml,qu2024omni}, frequency-domain artifact analysis~\citep{zhang2023perceptual} or or hybrid attention~\citep{xin2026hytrechybridtemporalawareattention,kong2025token}. Nevertheless, the resulting explanations are often difficult for humans to comprehend, as they lack clarity in natural language. This limitation is particularly evident in the video domain, where AI-generated content frequently exhibits obvious flaws, e.g., incoherent frame transitions and object inconsistency, that are easily noticed and reasoned about by humans~\citep{deng2024survey}. FakeClue~\citep{fakeclue_wen2025spot} introduces the use of vision-language models (VLMs) to provide interpretability for image-level detection, but it does not offer a unified framework that integrates both images and videos.

\subsection{Datasets for Synthetic Content Detection.}

Early datasets for synthetic content detection~\citep{zhang2025detecting,jiang2025revisiting,ma2025imdl,du2025forensichub,zhu2025does}, such as CNNSpot~\citep{cnnspot_wang2020cnn}, primarily collected fake images generated by GAN-based models~\citep{gan1_goodfellow2014generative,gan2_zhu2017unpaired,gan3_brock2018large}. However, with the advent of more advanced generative architectures like diffusion models~\citep{diff4_hertz2022prompt,diff5_nichol2021glide,ying2026beyond} and their variants, the authenticity of generated content has significantly increased, making it more challenging for detection models to discern. This has spurred the development of newer datasets, including ArtiFact~\citep{cao2024synartifact}, GenImage~\citep{genimage_2023}, GenImagePP~\citep{zhou2025breaking} and WildFake~\citep{wildfake_2025}. GenImage~\citep{genimage_2023} comprises images from the 1000 ImageNet~\citep{imagenet_ILSVRC15} categories, generated by eight state-of-the-art generators such as Stable Diffusion~\citep{rombach2022high} and Midjourney. Nevertheless, these datasets predominantly focus on image-based content. More recently, datasets emphasizing interpretability have also been introduced~\citep{zhao2026spinebench}. FakeClue~\citep{fakeclue_wen2025spot} contains a large amount of image data with explainability annotations but lacks video data. LOKI~\citep{loki_ye2024loki} offers data across 26 different categories and includes 18,000 distinct questions; however, its volume of multimodal data is relatively small and primarily suited for evaluation rather than comprehensive model training~\citep{ji2025zoom}. Therefore, a critical gap exists for a unified benchmark encompassing both image and video modalities to rigorously evaluate the performance of contemporary AIGC detectors.
\section{Methodology}

\subsection{Data Collection}

The \benchmark~dataset is curated from GenVideo~\citep{demamba}, LOKI~\citep{loki_ye2024loki}, FakeClue~\citep{fakeclue_wen2025spot}, WildFake~\citep{wildfake_2025}, Kinetics-400~\citep{kay2017kineticshumanactionvideo}, and MMFakeBench~\citep{liu2024mmfakebench}. Since \benchmark~is substantially larger than other datasets, we primarily detail our data processing procedures here.

\paragraph{Image/Video Corruption Filtering.} 
The integrity of images and videos is critical for MLLM training. We observed that the official GenVideo~\citep{demamba} repository contained a large number of unreadable videos. To ensure data quality, we employed the official Qwen2.5-VL~\citep{bai2025qwen2} I/O library to filter out corrupted files, discarding images and videos that failed to load.

\paragraph{Image/Video Resolution.} 
To maintain consistency, we only retained images whose resolutions fell within [3,136, 12,845,056] pixels, and videos whose per-frame resolutions were within [100,352, 602,112] pixels, with the additional constraint that total pixels did not exceed 19,267,968. Samples outside these ranges were discarded.

\paragraph{Near-Duplicate Image Filtering.} 
Selecting challenging and diverse samples is crucial for model robustness. We extracted visual features from images and video frames using CLIP~\citep{Radford2021LearningTV} and computed intra-class similarity scores. Images with excessively high similarity were removed, while videos were left unprocessed.
To comprehensively activate and enhance the model’s reasoning capabilities across the task, we propose a high-quality data construction pipeline that integrates explainable detection of AI-generated content.

\paragraph{Public Dataset Sources} 
The \benchmark~training set comprises approximately 106K image and video samples collected during the cold-start phase from diverse sources to ensure broad coverage of contemporary generative models. A significant portion was sourced from established public datasets, including GenVideo~\citep{demamba}, LOKI~\citep{loki_ye2024loki}, FakeClue~\citep{fakeclue_wen2025spot}, WildFake~\citep{wildfake_2025}, Kinetics-400~\citep{kay2017kineticshumanactionvideo}, MMFakeBench~\citep{liu2024mmfakebench}, GenImage~\citep{genimage_2023}, DiffusionForensics~\citep{dire_wang2023dire}. The cold-start phase aimed to enhance the granularity of interpretable content. To further improve the detector’s accuracy, an additional 87,500 image/video samples were stratified and sampled from multiple datasets for RLVR training, including MSRVTT~\citep{xu2016msrvtt} (9,000 videos), GenImage~\citep{genimage_2023} (32,000 images per category), CnnSpot~\citep{cnnspot_wang2020cnn} (19,000 samples), DigiFakeAV~\citep{liu2025faceswappingdiffusionbaseddigital} (2,000 samples), GenVideo~\citep{demamba} (Fake subset) (9,000 samples), DiffusionForensics~\citep{dire_wang2023dire} (4,500 samples), WildFake~\citep{wildfake_2025} (12,000 samples), and synthetic data generated by Qwen-Image~\citep{wu2025qwenimagetechnicalreport} (634 samples). In total, the dataset used for training comprises 106K samples.

\paragraph{Generated Sources}
To further enrich the dataset and better capture synthetic images prevalent in real-world scenarios, additional samples were generated using a diverse set of state-of-the-art image generation models. These models span multiple architectures and training paradigms, ensuring broad coverage of visual styles and generative patterns for robust evaluation. Specifically, extra sources includes samples generated by FLUX.1~\citep{labs2025flux1kontextflowmatching}, Lora Flux~\citep{labs2025flux1kontextflowmatching} and Qwen-Image~\citep{wu2025qwenimagetechnicalreport}.

\subsection{Data Annotation}

As illustrated in Table~\ref{tab:task_dimension}, ``file\_type'' indicates the modality of the input—either ``image'' or ``video'', and label represents the ground-truth label, assigned as either ``real'' or ``fake''.
The distilled explanations were further categorized along two major dimensions to facilitate structured analysis: \textbf{Spatial Features,} which comprises eight sub-dimensions and captures artifacts and inconsistencies observable within individual frames or static images. \textbf{Temporal Features,} which includes four sub-dimensions and describes anomalies associated with motion, temporal coherence, and cross-frame consistency. Since still images inherently lack temporal attributes, this category is exclusively applicable to video annotations. These categories were informed by established taxonomies of visual artifacts in generative content, as detailed in prior research~\citep{deng2024survey}.

\subsection{Comparison with Existing Datasets}
 
A comparative overview of \benchmark~and several existing AIGC detection datasets is provided in Table~\ref{tab:comparewithdataset}. Notably, \benchmark~offers unique advantages by integrating explainable annotations across both image and video modalities, addressing a significant gap in current resources.

\begin{table}[ht]
\caption{Comparison on the different datasets used in binary classification and interpretability tasks. Token lengths are computed using the GPT-4o tokenizer from the tiktoken library.}
\label{tab:comparewithdataset}
\centering
\resizebox{\linewidth}{!}{
\begin{tabular}{lccccc}
\toprule
\textbf{Dataset} & \textbf{Avg Token} & \textbf{Generator} & \textbf{Modality} & \multicolumn{2}{c}{\textbf{Dataset}} \\
\cmidrule(lr){5-6}
& \textbf{Lengths} & & & \textbf{fake} & \textbf{real} \\
\midrule
FakeBench~\citep{liu2024mmfakebench} & -   & 6     & Image         & 3K & 3K \\
VANE-Bench~\citep{bharadwaj2024vanebench} & 101    & 5     & Image*        & 2K & 1K \\
LOKI~\citep{loki_ye2024loki}  & 226.5 & $\sim$16 & Image+Video   & 3K & 0 \\
FakeClue~\citep{fakeclue_wen2025spot} & 119.6 & 26    & Image         & 68K & 36K \\
\benchmark~ & \textbf{451.7}  & \textgreater 30   & Image+Video   & \textbf{53K} & \textbf{78K} \\
\bottomrule
\end{tabular}}
\end{table}

 We identified a long-tail distribution in the distilled dataset, with a small portion of samples exhibiting token lengths concentrated at both extremes~\citep{xin2026hytrechybridtemporalawareattention}. To mitigate the influence of these outliers, we retained only the central 90\% of the distribution, filtering out the top and bottom 5\% of samples, thereby keeping data with token lengths between 60 and 800. As shown in Figure~\ref{fig:figure-data-staic} panel (a), Ivy-Fake exhibits the longest token length distribution, with an average token length of 451.7. Additionally, to conduct a fine-grained analysis of the semantic dimensions embedded in the explanatory language across different interpretability datasets, we employ Qwen3-32B~\citep{qwen3} for dimension extraction. To enhance the accuracy of the extraction, we embed supplementary contextual information from each dataset into the model input. For example, in LOKI~\citep{loki_ye2024loki}, the bounding box coordinates are wrapped with \textless bbox\textgreater{}\textless /bbox\textgreater{} tags, while metadata such as coordinates, titles, and fine-grained descriptions are also included as input to the model. Moreover, we set the Temperature to 0.8 and perform three extraction passes for each sample, retaining only the dimensions consistently identified across all runs. The final aggregated results are visualized in panel (b).

\begin{figure*}[htbp]
    \centering
    \includegraphics[width=0.95\textwidth]{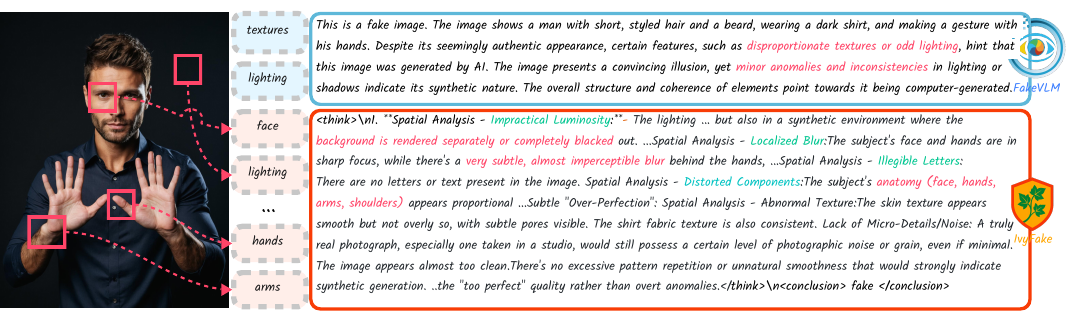}
    \caption{Comparison between \benchmark~and FakeVLM~\citep{fakeclue_wen2025spot} (NeurIPS 2025).}
\end{figure*}

\benchmark~outperforms FakeVLM in terms of interpretability. While FakeVLM only explains superficial factors such as lighting and texture, \benchmark~considers \textbf{multiple semantic dimensions}, including hair, skin, and background, leading to a more comprehensive understanding of forged content.

\subsection{Quality Control}


To mitigate the impact of hallucinations from Gemini 2.5 Pro during question generation, we sampled 8,000 cases for manual review. Each chain-of-thought (CoT) involving Gemini underwent at least two rounds of human verification. A total of ten users, including senior university students and regular participants, contributed to the verification process, which required approximately 1,000 hours in total. Each case was tested by at least two users to ensure robustness across different modalities of synthetic data. After this rigorous filtering, we retained 5,000 cases as the benchmark test set, consisting of 2,500 image samples and 2,500 video samples, with each modality containing 1,250 real and 1,250 fake instances.

\subsection{Method}

\begin{figure}[htbp]
    \centering
    \includegraphics[width=0.5\linewidth]{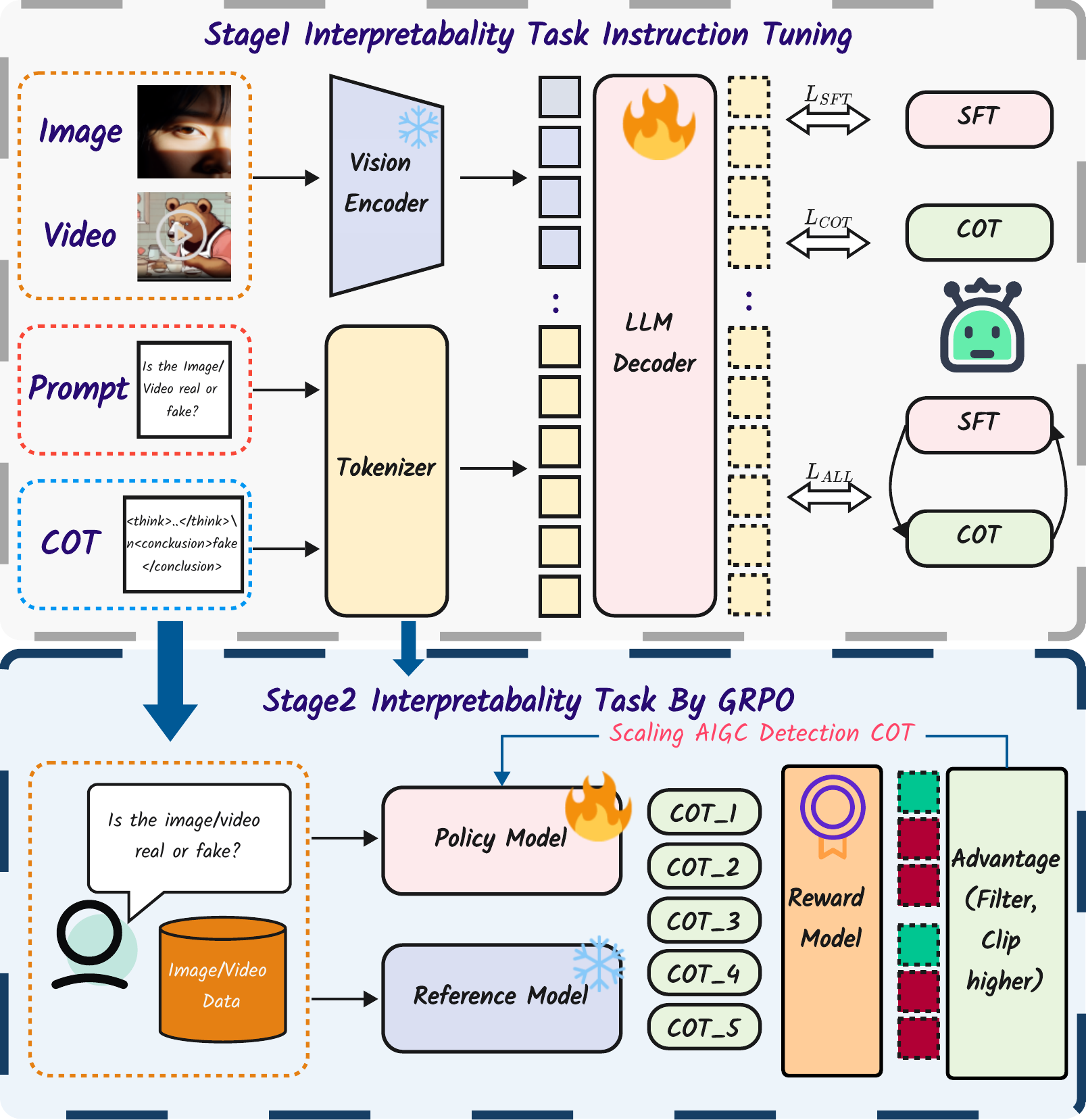}
    \caption{Overview of the three-stage training pipeline for \baseline.} 
    \label{fig:fig2} 
\end{figure}

Our preliminary investigations revealed that existing MLLMs exhibit inadequate performance on these tasks. To overcome this limitation, we propose \baseline, a multimodal large language model designed explicitly for robust and explainable AIGC detection. The overall training pipeline is illustrated in Figure~\ref{fig:fig2}.

\paragraph{Stage 1: Cold Start for Instruction-Driven Detection and Explainability.}
Due to limited reasoning ability, existing models struggle to detect AIGC content and generate reliable explanations. To address this, we inject Gemini-generated detection and explanatory COT into the training process, thereby improving the model’s fine-grained artifact perception and explanation quality.
This stage aims to initialize \baseline~with instruction following and explainability capabilities. The resulting model not only performs accurate AIGC detection but also generates coherent, human-understandable rationales for its classifications.

\begin{table*}[ht]
\caption{Performance comparison of models on image and video tasks. “Auto Metrics” include Acc, F1. “GPT Assisted” includes four subjective criteria: Comprehensiveness, Relevance, Detail, and Explanation. \textbf{Bold} indicates the best result, and \underline{underline} indicates the second best.}
\label{tab:model_performance}
\centering
\renewcommand{\arraystretch}{1.3}
\resizebox{\textwidth}{!}{
\begin{tabular}{l|cc|cc|cc}
\toprule
\multirow{3}{*}{\textbf{Model}} 
& \multicolumn{2}{c|}{\textbf{Image}} 
& \multicolumn{2}{c|}{\textbf{Video}} 
& \multicolumn{2}{c}{\textbf{Overall}} \\
\cmidrule(lr){2-3} \cmidrule(lr){4-5} \cmidrule(lr){6-7}
& \textbf{Auto Metrics} & \textbf{GPT Assisted}
& \textbf{Auto Metrics} & \textbf{GPT Assisted}
& \textbf{Auto Metrics} & \textbf{GPT Assisted} \\
\cmidrule(lr){2-3} \cmidrule(lr){4-5} \cmidrule(lr){6-7}
& Acc/F1/ROUGE-L/SIM & Com./Rel./Det./Exp.
& Acc/F1/ROUGE-L/SIM & Com./Rel./Det./Exp. 
& Acc/F1/ROUGE-L/SIM & Com./Rel./Det./Exp. \\
\midrule
\multicolumn{7}{c}{\textit{\textbf{Closed-source MLLMs}}} \\
\midrule
GPT-4o
& 0.725/0.723/0.108/0.525 & 2.34/3.20/2.04/3.26
& 0.448/0.579/0.072/0.451 & 1.79/2.35/1.67/2.40
& 0.587/0.663/0.090/0.488 & 2.07/2.78/1.85/2.83 \\
Gemini-2.5-Flash
& \underline{0.747}/\underline{0.737}/\underline{0.263}/\textbf{0.733} & \textbf{3.94}/\textbf{4.11}/\textbf{4.04}/\textbf{4.09}
& \underline{0.810}/\underline{0.811}/\underline{0.246}/\underline{0.723} & \textbf{4.00}/\textbf{4.37}/\textbf{4.03}/\textbf{4.36}
& \underline{0.779}/\underline{0.776}/\underline{0.254}/\textbf{0.728} & \textbf{3.97}/\textbf{4.24}/\textbf{4.04}/\textbf{4.22} \\
\midrule
\multicolumn{7}{c}{\textit{\textbf{Open-source MLLMs}}} \\
\midrule
\multicolumn{7}{{l}}{\textit{7B-Parameters Models}} \\
\midrule
InternVL3.5-8B         
& 0.605/0.602/0.194/0.680 & 2.83/3.49/2.69/3.32 & 0.574/0.588/0.188/0.664 & 2.75/3.35/2.68/3.28 & 0.589/0.596/0.191/0.672 & 2.79/3.42/2.69/3.30 \\
MiMo-VL-7B
& 0.662/0.637/0.121/0.593 & 1.99/2.80/1.85/2.89
& 0.778/0.783/0.112/0.580 & 2.04/2.91/1.90/3.19
& 0.720/0.715/0.116/0.586 & 2.01/2.86/1.87/3.04 \\
Qwen2.5-VL-7B
& 0.013/0.026/0.006/0.264 & 1.02/1.03/1.01/1.52
& 0.092/0.159/0.015/0.280 & 1.14/1.23/1.12/2.21
& 0.053/0.096/0.010/0.272 & 1.08/1.13/1.07/1.86 \\
LLaVA-OneVision-1.5-8B
& 0.500/0.333/0.080/0.499 & 1.51/2.62/1.49/2.38
& 0.500/0.333/0.068/0.481 & 1.49/2.26/1.37/2.19
& 0.500/0.333/0.074/0.490 & 1.50/2.44/1.43/2.28 \\
MiniCPM-V-4.5
& 0.666/0.680/0.169/0.637 & 3.20/3.93/3.06/3.60
& 0.491/0.505/0.152/0.627 & 2.83/3.66/2.76/3.36
& 0.579/0.610/0.161/0.632 & 3.01/3.80/2.91/3.48 \\
\midrule
\multicolumn{7}{l}{\textit{3B-Parameters Models}} \\
\midrule
Qwen2.5-VL-3B
& 0.641/0.612/0.023/0.391 & 1.19/1.33/1.18/3.28
& 0.689/0.686/0.017/0.381 & 1.42/1.56/1.40/3.68
& 0.665/0.652/0.020/0.386 & 1.31/1.45/1.29/3.48 \\
Gemma-3-4b-it
& 0.408/0.477/0.170/0.576 & 2.55/3.36/2.46/3.11
& 0.396/0.482/0.149/0.561 & 2.30/3.03/2.37/2.95
& 0.402/0.482/0.159/0.568 & 2.43/3.19/2.42/3.03 \\
InternVL3.5-2B
& 0.602/0.573/0.177/0.648 & 2.62/3.29/2.51/3.13
& 0.435/0.459/0.159/0.631 & 2.46/3.16/2.42/2.99
& 0.518/0.518/0.168/0.640 & 2.54/3.22/2.47/3.06 \\
InternVL3.5-4B
& 0.651/0.652/0.190/0.660 & 3.01/3.68/2.95/3.58
& 0.614/0.617/0.181/0.653 & 2.93/3.61/2.83/3.52
& 0.632/0.635/0.186/0.656 & 2.97/3.64/2.89/3.55 \\
\baseline
& \textbf{0.831}/\textbf{0.831}/\textbf{0.283}/\underline{0.714} & \underline{3.54}/\underline{4.04}/\underline{3.61}/\underline{3.85}
& \textbf{0.897}/\textbf{0.897}/\textbf{0.300}/\textbf{0.726} & \underline{3.72}/\underline{4.12}/\underline{3.75}/\underline{4.24}
& \textbf{0.864}/\textbf{0.864}/\textbf{0.291}/\underline{0.720} & \underline{3.63}/\underline{4.08}/\underline{3.68}/\underline{4.05}\\
\bottomrule
\end{tabular}}
\end{table*}

\paragraph{Stage 2: Sparse Rewards in Fine-grained Visual forgery Reasoning via Reinforcement Learning}

Although the fine-tuned model from Stage~1 demonstrates improved artifact awareness and interpretability, its generalization ability remains limited. To further enhance the model’s capacity for producing consistent and human-understandable explanations, we adopt the Group Relative Policy Optimization (GRPO) algorithm~\citep{shao2024deepseekmathpushinglimitsmathematical}. 
GRPO enables reinforcement learning~\citep{zhoubao-earlvr,zhao2026know} using only binary classification samples, achieving efficient and data-light optimization.

We construct binary real/fake pairs from multiple datasets, each containing both authentic and synthetic images and videos. \baseline~takes the visual input and extracts the text enclosed within the \verb|<conclusion>|... \verb|</conclusion>| tags, where the predicted label is either \textit{real} or \textit{fake}. The quantitative results of different training stages are presented in Table~\ref{tab:ablation_sft_rl}.

\subsection{GRPO}

Following DeepSeek-R1~\citep{deepseekr1_guo2025deepseek}  and TextShield-R1 \cite{qu2026textshield}, we adapt the Group Relative Policy Optimization (GRPO)~\citep{shao2024deepseekmathpushinglimitsmathematical}, an online RL algorithm designed to maximize the advantage of generated completions while constraining policy divergence from a reference model. We formalize our training process of \baseline~using GRPO below. Let $p$ denote a sampled prompt, and let ${o_{1}, o_{2}, . . . , o_{n}}$ be a group of completions generated by the current policy $\pi_{\theta}$. For each completion $G_{i}$, a reward $r_{i}$ is computed using a rule-based reward function. The group advantage for each completion is then calculated as:

\begin{equation}
\hat{A}_{i,t}=\frac{r_{i}-mean(r)}{std(r)}
\end{equation}

\begin{figure*}[t]
    \begin{equation}
\mathcal{L}_{\mathrm{GRPO}}(\theta)
= -\frac{1}{K}\sum_{k=1}^{K} A^{(k)}\,\ell^{(k)}(\theta)
\;+\; \beta\,\mathrm{KL}\!\left(\pi_{\theta}(\cdot|x)\,\|\,\pi_{\mathrm{ref}}(\cdot|x)\right).
\end{equation}
\end{figure*}

where $\beta$ is a coefficient that balances advantage maximization and KL regularization, and the clipping operator $clip(\dots, 1-\epsilon, 1+\epsilon)$ constrains the update magnitude. To regularize policy updates, we estimate the token-level Kullback-Leibler (KL) divergence between the current policy $\pi_{\theta}$ and a reference policy $\pi_{ref}$.

\begin{figure*}[t]
\begin{equation}
\mathbb{D}_{\text{KL}} \big[ \pi_\theta \big\| \pi_{\text{ref}} \big] = \frac{\pi_{\text{ref}}(o_{i,t} \mid p, o_{i,<t})}{\pi_\theta(o_{i,t} \mid p, o_{i,<t})} - \log \frac{\pi_{\text{ref}}(o_{i,t} \mid p, o_{i,<t})}{\pi_\theta(o_{i,t} \mid p, o_{i,<t})} - 1.
\label{fig:eq_kl}
\end{equation}
\end{figure*}

\subsection{Reward Model}
For effective RL, we employ a rule-based reward that consists of accuracy and format rewards. We introduce a solid accuracy reward
for AIGC Detection, which utilizes distinct functions to evaluate binary classification task. This allows for a more appropriate assessment based on the expected answer type.

\begin{itemize}
    \item \textbf{Accuracy Reward}: The accuracy reward assigns a score of \textbf{1} if the label in \texttt{<conclusion>} exactly matches the ground-truth classification \texttt{real} and \texttt{fake} and \textbf{0} otherwise.
    \item \textbf{Format Reward}: The format reward assigns a score of \textbf{1} if the output strictly follows the structural requirements by enclosing the reasoning within \texttt{<think></think>} tags and the final decision within \texttt{<conclusion></conclusion>} tags, and \textbf{0} otherwise.
\end{itemize}

\section{Experiments}

\subsection{Experimental Details}

\paragraph{Baselines} We primarily evaluate three closed-source models and three open-source models on our \benchmark. The closed-source models are GPT-4o~\citep{gpt4_achiam2023gpt} and Gemini2.5 Flash~\citep{genimi_team2023gemini}. For the open-source models, we select models of comparable size: InternVL3.5~\citep{chen2024internvl}, Qwen2.5-VL~\citep{bai2025qwen2}, MiMo-VL~\citep{coreteam2025mimovltechnicalreport}, LLaVA-OneVision-1.5~\citep{LLaVA-OneVision-1.5,lillava} and MiniCPM-V-4.5~\citep{yao2024minicpm,yu2025minicpmv45cookingefficient}. \baseline~denotes our fully RL-trained model after the GRPO optimization stage.

\paragraph{Evaluation} We report standard accuracy (Acc) and macro-averaged F1 score (F1) to assess the model’s ability to distinguish real from fake instances. For the reasoning task, we measure the similarity between the model’s reasoning process and the reference annotations using the ROUGE-L score~\citep{lin2004rouge} and BertScore~\citep{bert-score,he2021deberta}, which captures the longest common subsequence between predicted and reference texts, reflecting token-level overlap.  Since ROUGE-L may fail to fully capture the fidelity of reasoning steps, we adopt an LLM-as-a-judge evaluation paradigm~\citep{zheng2023judging}, following the FakeBench protocol~\citep{liu2024mmfakebench}, which assesses model responses along four dimensions: (1) Completeness: It reflects the extent to which the response fully addresses all aspects of the user’s question. More complete responses should incorporate information aligning well with the “golden clues” or reference answers. Incomplete or partially answered responses will receive lower scores. (2) Relevance: Measure how closely the content relates to the original annotation; (3) Level of Detail: Assess whether the response includes enough examples or elaborations; (4) Explanation: Verify the accuracy and consistency of explanations for any causes mentioned. Each response is scored using GPT-4o mini~\citep{gpt4_achiam2023gpt}, which instructs the model to act as an impartial judge and assign a score from 1 to 5.

\begin{figure*}[ht]
    \centering
    \includegraphics[width=0.95\textwidth]{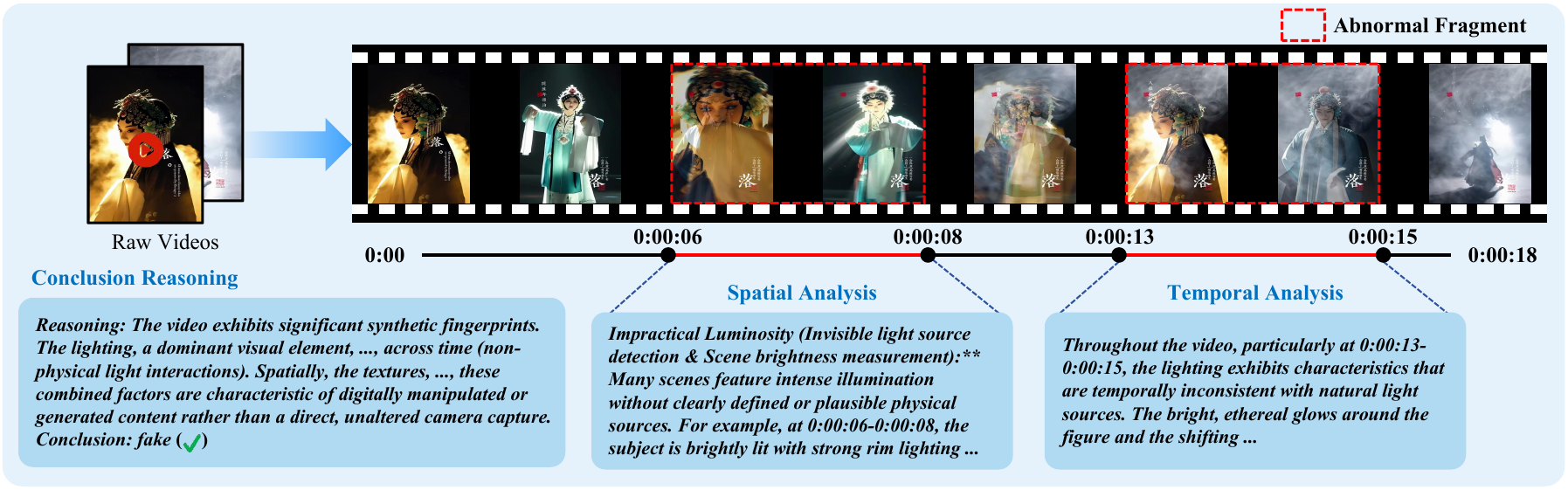}
    \caption{\baseline~performance on video detection.}
    \label{fig:case-ivyxd}
\end{figure*}

\paragraph{Implementation Details} Our model was trained using the AdamW~\citep{adamw_Loshchilov2017FixingWD} optimizer with a cosine learning rate schedule. During the SFT stage, the model was trained for one epoch with a batch size of 1, which ensures stable convergence on large-scale multimodal data. In the subsequent RL stage, we adopted the GRPO algorithm, training for one epoch with the same batch size and a learning rate of $1\times10^{-5}$. Within GRPO, the group size $n$ was set to 8, the warm-up ratio to 0.01, the temperature to 0.9. Training the SFT and RL stages required approximately 10 and 50 hours, respectively, on a system equipped with 32 A100 GPUs.

For video-based inputs, we configured the preprocessing pipeline to sample frames at a rate of 1 frame per second, with a maximum of 6 frames per clip to ensure temporal coherence while maintaining computational efficiency.
Across the training pipeline, the overall input resolution was constrained by a global upper bound of 6,422,528 pixels (MAX\_PIXELS), preventing overflow during multi-GPU parallel training and ensuring consistent input scaling across image and video modalities.

\subsection{Main Results}

\begin{table*}[!ht]
\centering
\caption{Comparison on the \textbf{Chameleon}~\citep{aide_yan2025a}. Accuracy (\%) of different detectors (rows) in detecting real and fake images. For each training dataset, the first row is overall accuracy, the second row is ``fake/real'' class accuracy.}
\resizebox{\textwidth}{!}{
\begin{tabular}{cccccccccccc}
\toprule
\textbf{CNNSpot} & \textbf{FreDect} & \textbf{Fusing} & \textbf{GramNet} & \textbf{LNP} & \textbf{UnivFD} & \textbf{DIRE} & \textbf{PatchCraft} & \textbf{NPR} & \textbf{AIDE} & \textbf{\baseline}\\
\midrule
60.89 & 57.22 & 57.09 & 60.95 & 58.52 & 60.42 & 59.71 & 56.32 & 58.13 & 65.77 & \textbf{73.17} \\
9.86/99.25 & 0.89/99.55 & 0.02/99.98 & 4.76/99.66 & 7.72/96.70 & 85.52/41.56 & 11.86/95.67 & 3.07/96.35 & 2.43/100.00 & 26.80/95.06 & 67.78/77.22\\
\bottomrule
\end{tabular}
}
\label{tab:chameleon}
\end{table*}
\begin{table*}[t]
\centering
\caption{Comparison on the \textbf{Genimage}~\citep{genimage_2023}. Accuracy (\%) of different detectors (rows) in detecting real and fake images from different generators (columns). The best result and the second-best result are marked in bold and underline, respectively. Results of other methods are reported from~\citep{aide_yan2025a}.}
\label{tab:genimage}
\resizebox{\linewidth}{!}{
\begin{tabular}{lccccccccccl}
\toprule
\textbf{Method} & \textbf{Midjourney} & \textbf{SD v1.4} & \textbf{SD v1.5} & \textbf{ADM} & \textbf{GLIDE} & \textbf{Wukong} & \textbf{VQDM} & \textbf{BigGAN} & \textbf{Mean} \\
\midrule
CNNSpot~\citep{cnnspot_wang2020cnn} & 52.80 & 96.30 & 95.90 & 50.10 & 39.80 & 78.60 & 53.40 & 46.80 & 64.21 \\
F3Net~\citep{qian2020thinking} & 50.10 & \textbf{99.90} & \textbf{99.90} & 49.90 & 50.00 & \textbf{99.90} & 49.90 & 49.90 & 68.69 \\
DIRE~\citep{dire_wang2023dire} & 60.20 & \textbf{99.90} & \underline{99.80} & 50.90 & 55.00 & \underline{99.20} & 50.10 & 50.20 & 70.66 \\
GenDet~\citep{zhu2023gendet} & 89.60 & 96.10 & 96.10 & 58.00 & 78.40 & 92.80 & 66.50 & 75.00 & 81.56 \\
PatchCraft~\citep{aigcdetectionbenchmark_zhong2023patchcraft} & 79.00 & 89.50 & 89.30 & 77.30 & 78.40 & 89.30 & 83.70 & 72.40 & 82.30 \\
FRIDA~\citep{bonechi2025thisfakedetectionsource} & 85.80 & 85.00 & 85.50 & 84.00 & 87.30 & 83.90 & 86.50 & 84.80 & 85.30 \\
AIDE~\citep{aide_yan2025a} & 79.38 & \underline{99.74} & 99.76 & 78.54 & 91.82 & 98.65 & 80.26 & 66.89 & 86.88 \\
DRCT~\citep{drct} & 91.50 & 95.00 & 94.40 & \underline{79.40} & 89.20 & 94.70 & 90.00 & 81.70 & 89.50 \\
Effort~\citep{effort} & 82.40 & 99.80 & 99.80 & 78.70 & \underline{93.30} & 97.40 & \underline{91.70} & \underline{77.60} & \underline{91.10} \\
\midrule
ThinkFake-7B~\citep{huang2025thinkfakereasoningmultimodallarge} & \underline{92.50} & 93.10 & 95.30 & 73.10 & 87.40 & 93.60 & 66.20 & 70.80 & 84.00 \\
\textbf{\baseline} & \textbf{96.41} & 97.07 & 97.14 & \textbf{95.59} & \textbf{95.95} & 96.92 & \textbf{96.03} & \textbf{95.22} & \textbf{96.32} \\
\bottomrule

\bottomrule

\end{tabular}
}
\end{table*}
\begin{table}[t]
\centering
\scriptsize
\caption{Comparisons to the \textbf{GenVideo}~\citep{demamba}. F1 score (F1), recall score (R) and average precision (AP) on the many-to-many generalization task. ``Demamba-XCLIP-FT'' is abbreviated as ``Demamba''. Results of other methods are reported from~\citep{demamba}.}
\label{tab:genvideo}
\resizebox{\linewidth}{!}{ 
\begin{tabular}{ll*{11}{c}c}
\toprule
Model & Metric & Sora & Morph & Gen2 & HotShot & Lavie & Show-1 & Moon & Crafter & Model & Wild & Avg. \\
& & & Studio & & & & & Valley & & Scope & Scrape & \\
\midrule
\multirow{2}{*}{F3Net (Image)} 
  & R  & 0.8393 & 0.9971 & 0.9862 & 0.7757 & 0.5700 & 0.3657 & 0.9952 & 0.9971 & 0.8943 & 0.7678 & 0.8188 \\
  & F1 & 0.5000 & 0.9406 & 0.9628 & 0.8169 & 0.6988 & 0.4904 & 0.9332 & 0.9688 & 0.8873 & 0.8251 & 0.8024 \\
\midrule
\multirow{2}{*}{NPR (Image)} 
  & R  & 0.9107 & 0.9957 & 0.9949 & 0.2429 & 0.8964 & 0.5771 & 0.9712 & 0.9986 & 0.9429 & 0.8780 & 0.8408 \\
  & F1 & 0.2786 & 0.8441 & 0.9131 & 0.3028 & 0.8627 & 0.5944 & 0.8170 & 0.9164 & 0.8184 & 0.8163 & 0.7164 \\
\midrule
\multirow{2}{*}{STIL (Video)} 
  & R  & 0.7857 & 0.9814 & 0.9804 & 0.7600 & 0.6179 & 0.5329 & 0.9936 & 0.9736 & 0.9457 & 0.6501 & 0.8222 \\
  & F1 & 0.3805 & 0.9068 & 0.9458 & 0.7824 & 0.7232 & 0.6217 & 0.9039 & 0.9433 & 0.8884 & 0.7267 & 0.7823 \\
\midrule
\multirow{2}{*}{DeMamba (Video)} 
  & R  & 0.9812 & 1.0000 & 0.9986 & 0.6543 & 0.9486 & 0.9886 & 1.0000 & 1.0000 & 0.9286 & 0.8909 & 0.9302 \\
  & F1 & 0.6407 & 0.9602 & 0.9790 & 0.7539 & 0.9537 & 0.9551 & 0.9557 & 0.9797 & 0.9240 & 0.9120 & 0.9020 \\
\midrule
\multirow{2}{*}{\baseline} 
  & R  & 0.7857 & 0.9371 & 0.9507 & 0.9443 & 0.9550 & 0.9643 & 0.9968 & 0.9857 & 0.8943 & 0.9461 & 0.9528 \\
  & F1 & 0.8800 & 0.9676 & 0.9747 & 0.9713 & 0.9770 & 0.9818 & 0.9984 & 0.9928 & 0.9442 & 0.9723 & 0.9526 \\ 
\bottomrule
\end{tabular}
} 
\end{table}

We perform extensive experiments to assess both detection and explanation capabilities. In particular, the proposed method is evaluated on the classification (real/fake) tasks and reasoning tasks for both image and video content using the proposed unified framework. For the classification task, we test our model on both image and video content to detect the synthesic content. 

From Table~\ref{tab:model_performance}, our method demonstrates robust detection and explanation capabilities across both image and video tasks. Overall, closed-source models maintain a clear advantage in classification accuracy and subjective explanation quality. Among them, Gemini-2.5-flash achieves the best performance on both automatic metrics and GPT-assisted evaluation (e.g., reaching Acc/F1 of 0.812/0.812 on video detection and an average explanation score above 4.0), reflecting strong overall detection and reasoning ability. In contrast, GPT-4o maintains relatively high classification accuracy.

For open-source models, most models in the 3B–7B parameter range exhibit relatively low scores on explanation dimensions ($< 3.5$), indicating limitations in generating reasoning chains and covering multi-dimensional details. Notably, Qwen2.5-VL-3B shows relatively stable performance in explanation relevance and detail (Det./Exp. $> 3.2$), suggesting that lightweight models still hold potential under specific designs.

Notably, Gemini-2.5-Flash achieves the highest scores across all explanation-related dimensions, with an average of 4.12, surpassing all other models. This advantage can be attributed to the fact that part of our training data was distilled from Gemini-2.5-Pro, a powerful multimodal model capable of handling complex reasoning and explanation tasks. In contrast, our \baseline~ranks second with an average explanation score of 3.85, demonstrating interpretability comparable to Gemini-2.5-Flash while achieving substantially higher detection accuracy (Acc/F1 = 0.864/0.864) than any other open-source model. Compared with Gemini-2.5-Flash (Acc/F1 = 0.780/0.776), \baseline~improves accuracy by over 10 percentage points, confirming its robustness in both synthetic image detection and interpretive reasoning tasks.


\subsection{Generalization Evaluation}

\paragraph{Evaluation on Classification Benchmarks.}
We conduct evaluations across multiple public leaderboards, including the image-based GenImage~\ref{tab:genimage} and the video-based GenVideo benchmark~\ref{tab:genvideo}.
Performance results of other competing detectors are taken from AIDE~\citep{aide_yan2025a} and Demamba~\citep{demamba}.
Our method achieves superior performance on all these leaderboards, demonstrating superior detection accuracy and robust cross-modal consistency.

\paragraph{Evaluation on Unseen Benchmarks.}
Beyond the datasets aligned with our training sets, we further evaluate the generalization ability of \baseline~on the Chameleon benchmark\cite{aide_yan2025a}, which lies outside the training data distribution, as shown in Table~\ref{tab:chameleon}.
The results confirm that our approach preserves strong generalization capability across unseen generative models and diverse data domains.

Compared to the original leaderboard SOTA method \textbf{AIDE}, our model achieves an overall accuracy of \textbf{73.17\%}. Notably, while the accuracy for the \textit{real} class decreases from 95.06\% to 77.22\%, the accuracy for the \textit{fake} class substantially increases from 26.80\% to 67.78\%.
This suggests that \textbf{AIDE} tends to overfit the real category, whereas our model achieves a more balanced and robust performance across both real and synthetic content.

\paragraph{Ablation Study.}

\begin{table}[t]
  \caption{
    Ablation study on different SFT and RL training settings. 
    \textit{Base Model} denotes the base model Qwen2.5-VL-3B. Evaluation metric: ACC / F1.
  }
  \label{tab:ablation_sft_rl}
  \centering
  \resizebox{0.6\columnwidth}{!}{%
    \begin{tabular}{c|cc|ccc}
      \toprule
      \multirow{2}{*}{\textbf{Model Name}} 
        & \multicolumn{2}{c|}{\textbf{Training Stage}} 
        & \multirow{2}{*}{\textbf{Chameleon (Image)}} 
        & \multirow{2}{*}{\textbf{GenVideo (Video)}} 
        & \multirow{2}{*}{\textbf{\benchmark~Test}} \\
      & \textbf{SFT} & \textbf{RL} & & & \\
      \midrule
      Base Model &  &  
        & 59.05 / 53.30 
        & 67.70 / 80.74 
        & 65.10 / 61.80 \\
       - & \checkmark &  
        & 40.31 / 37.99 
        & \underline{91.61} / \underline{91.55} 
        & 73.91 / 73.83 \\
       - &  & \checkmark 
        & \underline{72.72} / \underline{70.79} 
        & 78.19 / 77.62 
        & \underline{77.00} / \underline{76.66} \\
       - & \checkmark & \checkmark 
        & \textbf{73.17} / \textbf{73.14} 
        & \textbf{95.29} / \textbf{95.29} 
        & \textbf{86.40} / \textbf{86.30} \\
      \bottomrule
    \end{tabular}
  }
\end{table}

To validate the effectiveness of our training framework, we design four ablation variants, as shown in Table~\ref{tab:ablation_sft_rl}.
As shown, applying SFT alone results in only a slight performance drop on the out-of-domain benchmark Chameleon, while achieving consistent improvements across all other datasets. This suggests that the SFT stage primarily enhances the model’s fine-grained sensitivity to synthetic artifacts. In contrast, applying RL alone yields modest but consistent gains in classification accuracy across benchmarks. The combination of SFT and RL achieves the best balance between accuracy and interpretability.

\section{Conclusion}

We introduced \benchmark, the first unified, large-scale dataset for explainable AIGC detection across both images and videos, featuring over 106K richly annotated training samples and 5,000 evaluation examples with natural-language reasoning, and proposed Ivy Explainable Detector (\baseline), a MLLM that jointly detects and explains synthetic content. Our model sets superior benchmarks in AIGC detection and explainability, and our publicly released resources provide a robust foundation for transparent, trustworthy multimodal analysis. We believe that such fine-grained explainability data \benchmark~can catalyze new research directions for the AIGC detection community, enabling deeper understanding of synthetic content and more principled model development.

\noindent \textbf{Limitations}: Future work will concentrate on tailoring the benchmark toward domain-specific scenarios, including explainability-focused datasets for DeepFake and document manipulation analysis.
Nevertheless, our experiments suggest that the proposed approach exhibits strong scalability and generalization capabilities, indicating the potential for even better performance when applied to larger models such as 32B or 72B.

\noindent \textbf{Broader impacts}: The interpretability evaluation framework we propose is currently applied only to the field of Fake Image Detection~\citep{du2025forensichub}, which is strictly a binary classification task. In future work, our multi-dimensional evaluation approach can be extended to assess the interpretability of other tasks, such as chain-of-thought evaluation for Table Question Answering~\citep{jiang2025tabdsr}, question answering in medical scenarios~\citep{zhao2026spinebench,xin2024artificial,xin2026dualcpt}, GUI Agents~\citep{yuanse}, Document Generation \cite{ying2024dig,ying2024dig,ying2024fine} and video grounding interpretability~\citep{lin2025audio}.

\newpage

\bibliographystyle{assets/plainnat}
\bibliography{paper}

@String(IJCV = {Int. J. Comput. Vis.})

@String(CVPR= {IEEE Conf. Comput. Vis. Pattern Recog.})

@String(ICCV= {Int. Conf. Comput. Vis.})

@String(NIPS= {Adv. Neural Inform. Process. Syst.})

@String(ICASSP=	{ICASSP})

@String(ICLR = {Int. Conf. Learn. Represent.})

@String(AAAI = {AAAI})

@String(IJCV  = {IJCV})

@String(CVPR  = {CVPR})

@String(ICCV  = {ICCV})

@String(NIPS  = {NeurIPS})

@String(ICLR  = {ICLR})

@article{sora_videoworldsimulators2024,
  title={Video generation models as world simulators},
  author={Tim Brooks and Bill Peebles and Connor Holmes and Will DePue and Yufei Guo and Li Jing and David Schnurr and Joe Taylor and Troy Luhman and Eric Luhman and Clarence Ng and Ricky Wang and Aditya Ramesh},
  year={2024},
  url={https://openai.com/research/video-generation-models-as-world-simulators},
  journal={arXiv preprint arXiv:2405.19707},
}

@article{dalle_3_betker2023improving,
  title={Improving image generation with better captions},
  author={Betker, James and Goh, Gabriel and Jing, Li and Brooks, Tim and Wang, Jianfeng and Li, Linjie and Ouyang, Long and Zhuang, Juntang and Lee, Joyce and Guo, Yufei and others},
  journal={Computer Science},
  volume={2},
  number={3},
  pages={8},
  year={2023}
}

@article{imagegen_saharia2022photorealistic,
  title={Photorealistic text-to-image diffusion models with deep language understanding},
  author={Saharia, Chitwan and Chan, William and Saxena, Saurabh and Li, Lala and Whang, Jay and Denton, Emily L and Ghasemipour, Kamyar and Gontijo Lopes, Raphael and Karagol Ayan, Burcu and Salimans, Tim and others},
  journal={Advances in neural information processing systems},
  volume={35},
  pages={36479--36494},
  year={2022}
}

@inproceedings{rombach2022high,
  title={High-resolution image synthesis with latent diffusion models},
  author={Rombach, Robin and Blattmann, Andreas and Lorenz, Dominik and Esser, Patrick and Ommer, Bj{\"o}rn},
  booktitle={Proceedings of the IEEE/CVF conference on computer vision and pattern recognition},
  pages={10684--10695},
  year={2022}
}

@article{Qureshi2024DeepfakeFA,
  title={Deepfake forensics: a survey of digital forensic methods for multimodal deepfake identification on social media},
  author={Shavez Mushtaq Qureshi and Atif Saeed and Sultan H. Almotiri and Farooq Ahmad and Mohammed A. Al Ghamdi},
  journal={PeerJ Computer Science},
  year={2024},
  volume={10},
  url={https://api.semanticscholar.org/CorpusID:270088699}
}

@inproceedings{10.5555/3294996.3295110,
author = {Steinhardt, Jacob and Koh, Pang Wei and Liang, Percy},
title = {Certified defenses for data poisoning attacks},
year = {2017},
isbn = {9781510860964},
publisher = {Curran Associates Inc.},
address = {Red Hook, NY, USA},
booktitle = {Proceedings of the 31st International Conference on Neural Information Processing Systems},
pages = {3520–3532},
numpages = {13},
location = {Long Beach, California, USA},
series = {NIPS'17}
}

@article{demamba,
      title={DeMamba: AI-Generated Video Detection on Million-Scale GenVideo Benchmark},
      author={Haoxing Chen and Yan Hong and Zizheng Huang and Zhuoer Xu and Zhangxuan Gu and Yaohui Li and Jun Lan and Huijia Zhu and Jianfu Zhang and Weiqiang Wang and Huaxiong Li},
      journal={arXiv preprint arXiv:2405.19707},
      year={2024}
}

@article{wildfake_2025,
    title={WildFake: A Large-Scale and Hierarchical Dataset for AI-Generated Images Detection},
    volume={39},
    url={https://ojs.aaai.org/index.php/AAAI/article/view/32363},
    DOI={10.1609/aaai.v39i4.32363},
    number={4},
    journal={Proceedings of the AAAI Conference on Artificial Intelligence},
    author={Hong, Yan and Feng, Jianming and Chen, Haoxing and Lan, Jun and Zhu, Huijia and Wang, Weiqiang and Zhang, Jianfu},
    year={2025},
    month={Apr.},
    pages={3500-3508}
}

@inproceedings{aigvdet_bai2024ai,
  title={AI-Generated Video Detection via Spatial-Temporal Anomaly Learning},
  author={Bai, Jianfa and Lin, Man and Cao, Gang and Lou, Zijie},
  booktitle={Chinese Conference on Pattern Recognition and Computer Vision (PRCV)},
  pages={460--470},
  year={2024},
  organization={Springer}
}

@article{loki_ye2024loki,
  title={LOKI: A Comprehensive Synthetic Data Detection Benchmark using Large Multimodal Models},
  author={Ye, Junyan and Zhou, Baichuan and Huang, Zilong and Zhang, Junan and Bai, Tianyi and Kang, Hengrui and He, Jun and Lin, Honglin and Wang, Zihao and Wu, Tong and others},
  journal={ICLR},
  year={2025}
}

@inproceedings{
    aide_yan2025a,
    title={A Sanity Check for {AI}-generated Image Detection},
    author={Shilin Yan and Ouxiang Li and Jiayin Cai and Yanbin Hao and Xiaolong Jiang and Yao Hu and Weidi Xie},
    booktitle={The Thirteenth International Conference on Learning Representations},
    year={2025},
    url={https://openreview.net/forum?id=ODRHZrkOQM}
}

@misc{labs2025flux1kontextflowmatching,
      title={FLUX.1 Kontext: Flow Matching for In-Context Image Generation and Editing in Latent Space},
      author={Black Forest Labs and Stephen Batifol and Andreas Blattmann and Frederic Boesel and Saksham Consul and Cyril Diagne and Tim Dockhorn and Jack English and Zion English and Patrick Esser and Sumith Kulal and Kyle Lacey and Yam Levi and Cheng Li and Dominik Lorenz and Jonas Müller and Dustin Podell and Robin Rombach and Harry Saini and Axel Sauer and Luke Smith},
      year={2025},
      eprint={2506.15742},
      archivePrefix={arXiv},
      primaryClass={cs.GR},
      url={https://arxiv.org/abs/2506.15742},
}

@article{svd_blattmann2023stable,
  title={Stable video diffusion: Scaling latent video diffusion models to large datasets},
  author={Blattmann, Andreas and Dockhorn, Tim and Kulal, Sumith and Mendelevitch, Daniel and Kilian, Maciej and Lorenz, Dominik and Levi, Yam and English, Zion and Voleti, Vikram and Letts, Adam and others},
  journal={arXiv preprint arXiv:2311.15127},
  year={2023}
}

@article{deepseekr1_guo2025deepseek,
  title={Deepseek-r1: Incentivizing reasoning capability in llms via reinforcement learning},
  author={Guo, Daya and Yang, Dejian and Zhang, Haowei and Song, Junxiao and Zhang, Ruoyu and Xu, Runxin and Zhu, Qihao and Ma, Shirong and Wang, Peiyi and Bi, Xiao and others},
  journal={arXiv preprint arXiv:2501.12948},
  year={2025}
}

@misc{shao2024deepseekmathpushinglimitsmathematical,
      title={DeepSeekMath: Pushing the Limits of Mathematical Reasoning in Open Language Models}, 
      author={Zhihong Shao and Peiyi Wang and Qihao Zhu and Runxin Xu and Junxiao Song and Xiao Bi and Haowei Zhang and Mingchuan Zhang and Y. K. Li and Y. Wu and Daya Guo},
      year={2024},
      eprint={2402.03300},
      archivePrefix={arXiv},
      primaryClass={cs.CL},
      url={https://arxiv.org/abs/2402.03300}, 
}

@inproceedings{genimage_2023,
    author = {Zhu, Mingjian and Chen, Hanting and Yan, Qiangyu and Huang, Xudong and Lin, Guanyu and Li, Wei and Tu, Zhijun and Hu, Hailin and Hu, Jie and Wang, Yunhe},
    title = {GenImage: a million-scale benchmark for detecting AI-generated image},
    year = {2023},
    publisher = {Curran Associates Inc.},
    address = {Red Hook, NY, USA},
    booktitle = {NeurIPS},
    articleno = {3398},
    numpages = {12},
    location = {New Orleans, LA, USA},
    series = {NIPS '23}
}

@inproceedings{cnnspot_wang2020cnn,
  title={CNN-generated images are surprisingly easy to spot... for now},
  author={Wang, Sheng-Yu and Wang, Oliver and Zhang, Richard and Owens, Andrew and Efros, Alexei A},
  booktitle={ICCV},
  pages={8695--8704},
  year={2020}
}

@inproceedings{qian2020thinking,
  title={Thinking in frequency: Face forgery detection by mining frequency-aware clues},
  author={Qian, Yuyang and Yin, Guojun and Sheng, Lu and Chen, Zixuan and Shao, Jing},
  booktitle={European conference on computer vision},
  pages={86--103},
  year={2020},
  organization={Springer}
}

@inproceedings{dire_wang2023dire,
  title={Dire for diffusion-generated image detection},
  author={Wang, Zhendong and Bao, Jianmin and Zhou, Wengang and Wang, Weilun and Hu, Hezhen and Chen, Hong and Li, Houqiang},
  booktitle={ICCV},
  pages={22445--22455},
  year={2023}
}

@article{zhu2023gendet,
  title={Gendet: Towards good generalizations for ai-generated image detection},
  author={Zhu, Mingjian and Chen, Hanting and Huang, Mouxiao and Li, Wei and Hu, Hailin and Hu, Jie and Wang, Yunhe},
  journal={arXiv preprint arXiv:2312.08880},
  year={2023}
}

@article{aigcdetectionbenchmark_zhong2023patchcraft,
  title={Patchcraft: Exploring texture patch for efficient ai-generated image detection},
  author={Zhong, Nan and Xu, Yiran and Li, Sheng and Qian, Zhenxing and Zhang, Xinpeng},
  journal={arXiv preprint arXiv:2311.12397},
  year={2023}
}

@inproceedings{liu2024mmfakebench,
  title={MMFakeBench: A Mixed-Source Multimodal Misinformation Detection Benchmark for LVLMs},
  author={Liu, Xuannan and Li, Zekun and Li, Peipei and Huang, Huaibo and Xia, Shuhan and Cui, Xing and Huang, Linzhi and Deng, Weihong and He, Zhaofeng},
  booktitle={ICLR},
  year={2025}
}

@article{fakeclue_wen2025spot,
  title={Spot the fake: Large multimodal model-based synthetic image detection with artifact explanation},
  author={Wen, Siwei and Ye, Junyan and Feng, Peilin and Kang, Hengrui and Wen, Zichen and Chen, Yize and Wu, Jiang and Wu, Wenjun and He, Conghui and Li, Weijia},
  journal={NeurIPS},
  year={2025}
}

@misc{kay2017kineticshumanactionvideo,
      title={The Kinetics Human Action Video Dataset}, 
      author={Will Kay and Joao Carreira and Karen Simonyan and Brian Zhang and Chloe Hillier and Sudheendra Vijayanarasimhan and Fabio Viola and Tim Green and Trevor Back and Paul Natsev and Mustafa Suleyman and Andrew Zisserman},
      year={2017},
      eprint={1705.06950},
      archivePrefix={arXiv},
      primaryClass={cs.CV},
      url={https://arxiv.org/abs/1705.06950}, 
}

@article{bai2025qwen2,
  title={Qwen2.5-vl technical report},
  author={Bai, Shuai and Chen, Keqin and Liu, Xuejing and Wang, Jialin and Ge, Wenbin and Song, Sibo and Dang, Kai and Wang, Peng and Wang, Shijie and Tang, Jun and others},
  journal={arXiv preprint arXiv:2502.13923},
  year={2025}
}

@article{genimi_team2023gemini,
  title={Gemini: a family of highly capable multimodal models},
  author={Team, Gemini and Anil, Rohan and Borgeaud, Sebastian and Alayrac, Jean-Baptiste and Yu, Jiahui and Soricut, Radu and Schalkwyk, Johan and Dai, Andrew M and Hauth, Anja and Millican, Katie and others},
  journal={arXiv preprint arXiv:2312.11805},
  year={2023}
}

@misc{bharadwaj2024vanebench,
      title={VANE-Bench: Video Anomaly Evaluation Benchmark for Conversational LMMs}, 
      author={Rohit Bharadwaj and Hanan Gani and Muzammal Naseer and Fahad Shahbaz Khan and Salman Khan},
      year={2024},
      eprint={2406.10326},
      archivePrefix={arXiv}
}

@article{qwen3,
    title={Qwen3 Technical Report}, 
    author={An Yang and Anfeng Li and Baosong Yang and Beichen Zhang and Binyuan Hui and Bo Zheng and Bowen Yu and Chang Gao and Chengen Huang and Chenxu Lv and Chujie Zheng and Dayiheng Liu and Fan Zhou and Fei Huang and Feng Hu and Hao Ge and Haoran Wei and Huan Lin and Jialong Tang and Jian Yang and Jianhong Tu and Jianwei Zhang and Jianxin Yang and Jiaxi Yang and Jing Zhou and Jingren Zhou and Junyang Lin and Kai Dang and Keqin Bao and Kexin Yang and Le Yu and Lianghao Deng and Mei Li and Mingfeng Xue and Mingze Li and Pei Zhang and Peng Wang and Qin Zhu and Rui Men and Ruize Gao and Shixuan Liu and Shuang Luo and Tianhao Li and Tianyi Tang and Wenbiao Yin and Xingzhang Ren and Xinyu Wang and Xinyu Zhang and Xuancheng Ren and Yang Fan and Yang Su and Yichang Zhang and Yinger Zhang and Yu Wan and Yuqiong Liu and Zekun Wang and Zeyu Cui and Zhenru Zhang and Zhipeng Zhou and Zihan Qiu},
    journal = {arXiv preprint arXiv:2505.09388},
    year={2025}
}

@article{deng2024survey,
  title={A survey of defenses against ai-generated visual media: Detection, disruption, and authentication},
  author={Deng, Jingyi and Lin, Chenhao and Zhao, Zhengyu and Liu, Shuai and Wang, Qian and Shen, Chao},
  journal={arXiv preprint arXiv:2407.10575},
  year={2024}
}

@inproceedings{Radford2021LearningTV,
  title={Learning Transferable Visual Models From Natural Language Supervision},
  author={Alec Radford and Jong Wook Kim and Chris Hallacy and Aditya Ramesh and Gabriel Goh and Sandhini Agarwal and Girish Sastry and Amanda Askell and Pamela Mishkin and Jack Clark and Gretchen Krueger and Ilya Sutskever},
  booktitle={International Conference on Machine Learning},
  year={2021},
  url={https://api.semanticscholar.org/CorpusID:231591445}
}

@article{gpt4_achiam2023gpt,
  title={Gpt-4 technical report},
  author={Achiam, Josh and Adler, Steven and Agarwal, Sandhini and Ahmad, Lama and Akkaya, Ilge and Aleman, Florencia Leoni and Almeida, Diogo and Altenschmidt, Janko and Altman, Sam and Anadkat, Shyamal and others},
  journal={arXiv preprint arXiv:2303.08774},
  year={2023}
}

@inproceedings{chen2024internvl,
title={Internvl: Scaling up vision foundation models and aligning for generic visual-linguistic tasks},
author={Chen, Zhe and Wu, Jiannan and Wang, Wenhai and Su, Weijie and Chen, Guo and Xing, Sen and Zhong, Muyan and Zhang, Qinglong and Zhu, Xizhou and Lu, Lewei and others},
booktitle={Proceedings of the IEEE/CVF Conference on Computer Vision and Pattern Recognition},
pages={24185--24198},
year={2024}
}

@misc{coreteam2025mimovltechnicalreport,
      title={MiMo-VL Technical Report}, 
      author={Core Team and Zihao Yue and Zhenru Lin and Yifan Song and Weikun Wang and Shuhuai Ren and Shuhao Gu and Shicheng Li and Peidian Li and Liang Zhao and Lei Li and Kainan Bao and Hao Tian and Hailin Zhang and Gang Wang and Dawei Zhu and Cici and Chenhong He and Bowen Ye and Bowen Shen and Zihan Zhang and Zihan Jiang and Zhixian Zheng and Zhichao Song and Zhenbo Luo and Yue Yu and Yudong Wang and Yuanyuan Tian and Yu Tu and Yihan Yan and Yi Huang and Xu Wang and Xinzhe Xu and Xingchen Song and Xing Zhang and Xing Yong and Xin Zhang and Xiangwei Deng and Wenyu Yang and Wenhan Ma and Weiwei Lv and Weiji Zhuang and Wei Liu and Sirui Deng and Shuo Liu and Shimao Chen and Shihua Yu and Shaohui Liu and Shande Wang and Rui Ma and Qiantong Wang and Peng Wang and Nuo Chen and Menghang Zhu and Kangyang Zhou and Kang Zhou and Kai Fang and Jun Shi and Jinhao Dong and Jiebao Xiao and Jiaming Xu and Huaqiu Liu and Hongshen Xu and Heng Qu and Haochen Zhao and Hanglong Lv and Guoan Wang and Duo Zhang and Dong Zhang and Di Zhang and Chong Ma and Chang Liu and Can Cai and Bingquan Xia},
      year={2025},
      eprint={2506.03569},
      archivePrefix={arXiv},
      primaryClass={cs.CL},
      url={https://arxiv.org/abs/2506.03569}, 
}

@inproceedings{LLaVA-OneVision-1.5,
  title={LLaVA-OneVision-1.5: Fully Open Framework for Democratized Multimodal Training},
  author={LLaVA Community Contributors},
  booktitle={arxiv},  
  year={2025}
 }

@article{lillava,
  title={LLaVA-OneVision: Easy Visual Task Transfer},
  author={Li, Bo and Zhang, Yuanhan and Guo, Dong and Zhang, Renrui and Li, Feng and Zhang, Hao and Zhang, Kaichen and Zhang, Peiyuan and Li, Yanwei and Liu, Ziwei and Li, Chunyuan},
  journal={Transactions on Machine Learning Research},
  year={2024}
}

@inproceedings{lin2004rouge,
  title={Rouge: A package for automatic evaluation of summaries},
  author={Lin, Chin-Yew},
  booktitle={Text summarization branches out},
  pages={74--81},
  year={2004}
}

@inproceedings{bert-score,
  title={BERTScore: Evaluating Text Generation with BERT},
  author={Tianyi Zhang and Varsha Kishore and Felix Wu and Kilian Q. Weinberger and Yoav Artzi},
  booktitle={International Conference on Learning Representations},
  year={2020},
  url={https://openreview.net/forum?id=SkeHuCVFDr}
}

@inproceedings{
he2021deberta,
title={DEBERTA: DECODING-ENHANCED BERT WITH DISENTANGLED ATTENTION},
author={Pengcheng He and Xiaodong Liu and Jianfeng Gao and Weizhu Chen},
booktitle={International Conference on Learning Representations},
year={2021},
url={https://openreview.net/forum?id=XPZIaotutsD}
}

@article{zheng2023judging,
  title={Judging llm-as-a-judge with mt-bench and chatbot arena},
  author={Zheng, Lianmin and Chiang, Wei-Lin and Sheng, Ying and Zhuang, Siyuan and Wu, Zhanghao and Zhuang, Yonghao and Lin, Zi and Li, Zhuohan and Li, Dacheng and Xing, Eric and others},
  journal={Advances in Neural Information Processing Systems},
  volume={36},
  pages={46595--46623},
  year={2023}
}

@article{adamw_Loshchilov2017FixingWD,
  title={Fixing Weight Decay Regularization in Adam},
  author={Ilya Loshchilov and Frank Hutter},
  journal={ArXiv},
  year={2017},
  volume={abs/1711.05101},
  url={https://api.semanticscholar.org/CorpusID:3312944}
}

@article{cao2024synartifact,
  title={Synartifact: Classifying and alleviating artifacts in synthetic images via vision-language model},
  author={Cao, Bin and Yuan, Jianhao and Liu, Yexin and Li, Jian and Sun, Shuyang and Liu, Jing and Zhao, Bo},
  journal={arXiv preprint arXiv:2402.18068},
  year={2024}
}

@article{keita2025bi,
  title={Bi-LORA: A Vision-Language Approach for Synthetic Image Detection},
  author={Keita, Mamadou and Hamidouche, Wassim and Bougueffa Eutamene, Hessen and Taleb-Ahmed, Abdelmalik and Camacho, David and Hadid, Abdenour},
  journal={Expert Systems},
  volume={42},
  number={2},
  pages={e13829},
  year={2025},
  publisher={Wiley Online Library}
}

@inproceedings{dong2022explaining,
  title={Explaining deepfake detection by analysing image matching},
  author={Dong, Shichao and Wang, Jin and Liang, Jiajun and Fan, Haoqiang and Ji, Renhe},
  booktitle={European conference on computer vision},
  pages={18--35},
  year={2022},
  organization={Springer}
}

@inproceedings{zhang2023perceptual,
  title={Perceptual artifacts localization for image synthesis tasks},
  author={Zhang, Lingzhi and Xu, Zhengjie and Barnes, Connelly and Zhou, Yuqian and Liu, Qing and Zhang, He and Amirghodsi, Sohrab and Lin, Zhe and Shechtman, Eli and Shi, Jianbo},
  booktitle={ICCV},
  pages={7579--7590},
  year={2023}
}

@article{gan1_goodfellow2014generative,
  title={Generative adversarial nets},
  author={Goodfellow, Ian J and Pouget-Abadie, Jean and Mirza, Mehdi and Xu, Bing and Warde-Farley, David and Ozair, Sherjil and Courville, Aaron and Bengio, Yoshua},
  journal={Advances in neural information processing systems},
  volume={27},
  year={2014}
}

@inproceedings{gan2_zhu2017unpaired,
  title={Unpaired image-to-image translation using cycle-consistent adversarial networks},
  author={Zhu, Jun-Yan and Park, Taesung and Isola, Phillip and Efros, Alexei A},
  booktitle={Proceedings of the IEEE international conference on computer vision},
  pages={2223--2232},
  year={2017}
}

@article{gan3_brock2018large,
  title={Large scale GAN training for high fidelity natural image synthesis},
  author={Brock, Andrew and Donahue, Jeff and Simonyan, Karen},
  journal={arXiv preprint arXiv:1809.11096},
  year={2018}
}

@article{diff1_ho2020denoising,
  title={Denoising diffusion probabilistic models},
  author={Ho, Jonathan and Jain, Ajay and Abbeel, Pieter},
  journal={Advances in neural information processing systems},
  volume={33},
  pages={6840--6851},
  year={2020}
}

@article{diff2_dhariwal2021diffusion,
  title={Diffusion models beat gans on image synthesis},
  author={Dhariwal, Prafulla and Nichol, Alexander},
  journal={Advances in neural information processing systems},
  volume={34},
  pages={8780--8794},
  year={2021}
}

@article{diff4_hertz2022prompt,
  title={Prompt-to-prompt image editing with cross attention control},
  author={Hertz, Amir and Mokady, Ron and Tenenbaum, Jay and Aberman, Kfir and Pritch, Yael and Cohen-Or, Daniel},
  journal={arXiv preprint arXiv:2208.01626},
  year={2022}
}

@article{diff5_nichol2021glide,
  title={Glide: Towards photorealistic image generation and editing with text-guided diffusion models},
  author={Nichol, Alex and Dhariwal, Prafulla and Ramesh, Aditya and Shyam, Pranav and Mishkin, Pamela and McGrew, Bob and Sutskever, Ilya and Chen, Mark},
  journal={arXiv preprint arXiv:2112.10741},
  year={2021}
}

@article{imagenet_ILSVRC15,
Author = {Olga Russakovsky and Jia Deng and Hao Su and Jonathan Krause and Sanjeev Satheesh and Sean Ma and Zhiheng Huang and Andrej Karpathy and Aditya Khosla and Michael Bernstein and Alexander C. Berg and Li Fei-Fei},
Title = {{ImageNet Large Scale Visual Recognition Challenge}},
Year = {2015},
journal   = {International Journal of Computer Vision (IJCV)},
doi = {10.1007/s11263-015-0816-y},
volume={115},
number={3},
pages={211-252}
}

@misc{wu2025qwenimagetechnicalreport,
      title={Qwen-Image Technical Report}, 
      author={Chenfei Wu and Jiahao Li and Jingren Zhou and Junyang Lin and Kaiyuan Gao and Kun Yan and Sheng-ming Yin and Shuai Bai and Xiao Xu and Yilei Chen and Yuxiang Chen and Zecheng Tang and Zekai Zhang and Zhengyi Wang and An Yang and Bowen Yu and Chen Cheng and Dayiheng Liu and Deqing Li and Hang Zhang and Hao Meng and Hu Wei and Jingyuan Ni and Kai Chen and Kuan Cao and Liang Peng and Lin Qu and Minggang Wu and Peng Wang and Shuting Yu and Tingkun Wen and Wensen Feng and Xiaoxiao Xu and Yi Wang and Yichang Zhang and Yongqiang Zhu and Yujia Wu and Yuxuan Cai and Zenan Liu},
      year={2025},
      eprint={2508.02324},
      archivePrefix={arXiv},
      primaryClass={cs.CV},
      url={https://arxiv.org/abs/2508.02324}, 
}

@misc{liu2025faceswappingdiffusionbaseddigital,
      title={Beyond Face Swapping: A Diffusion-Based Digital Human Benchmark for Multimodal Deepfake Detection}, 
      author={Jiaxin Liu and Jia Wang and Saihui Hou and Min Ren and Huijia Wu and Long Ma and Renwang Pei and Zhaofeng He},
      year={2025},
      eprint={2505.16512},
      archivePrefix={arXiv},
      primaryClass={cs.CV},
      url={https://arxiv.org/abs/2505.16512}, 
}

@inproceedings{xu2016msrvtt,
  title={Msr-vtt: A large video description dataset for bridging video and language},
  author={Xu, Jun and Mei, Tao and Yao, Ting and Rui, Yong},
  booktitle={CVPR},
  year={2016}
}

@misc{yu2025minicpmv45cookingefficient,
      title={MiniCPM-V 4.5: Cooking Efficient MLLMs via Architecture, Data, and Training Recipe}, 
      author={Tianyu Yu and Zefan Wang and Chongyi Wang and Fuwei Huang and Wenshuo Ma and Zhihui He and Tianchi Cai and Weize Chen and Yuxiang Huang and Yuanqian Zhao and Bokai Xu and Junbo Cui and Yingjing Xu and Liqing Ruan and Luoyuan Zhang and Hanyu Liu and Jingkun Tang and Hongyuan Liu and Qining Guo and Wenhao Hu and Bingxiang He and Jie Zhou and Jie Cai and Ji Qi and Zonghao Guo and Chi Chen and Guoyang Zeng and Yuxuan Li and Ganqu Cui and Ning Ding and Xu Han and Yuan Yao and Zhiyuan Liu and Maosong Sun},
      year={2025},
      eprint={2509.18154},
      archivePrefix={arXiv},
      primaryClass={cs.LG},
      url={https://arxiv.org/abs/2509.18154}, 
}

@article{yao2024minicpm,
  title={MiniCPM-V: A GPT-4V Level MLLM on Your Phone},
  author={Yao, Yuan and Yu, Tianyu and Zhang, Ao and Wang, Chongyi and Cui, Junbo and Zhu, Hongji and Cai, Tianchi and Li, Haoyu and Zhao, Weilin and He, Zhihui and others},
  journal={Nat Commun 16, 5509 (2025)},
  year={2025}
}

@misc{huang2025thinkfakereasoningmultimodallarge,
      title={ThinkFake: Reasoning in Multimodal Large Language Models for AI-Generated Image Detection}, 
      author={Tai-Ming Huang and Wei-Tung Lin and Kai-Lung Hua and Wen-Huang Cheng and Junichi Yamagishi and Jun-Cheng Chen},
      year={2025},
      eprint={2509.19841},
      archivePrefix={arXiv},
      primaryClass={cs.CV},
      url={https://arxiv.org/abs/2509.19841}, 
}

@misc{bonechi2025thisfakedetectionsource,
      title={Who Made This? Fake Detection and Source Attribution with Diffusion Features}, 
      author={Simone Bonechi and Paolo Andreini and Barbara Toniella Corradini},
      year={2025},
      eprint={2510.27602},
      archivePrefix={arXiv},
      primaryClass={cs.CV},
      url={https://arxiv.org/abs/2510.27602}, 
}

@inproceedings{drct,
author = {Chen, Baoying and Zeng, Jishen and Yang, Jianquan and Yang, Rui},
title = {DRCT: diffusion reconstruction contrastive training towards universal detection of diffusion generated images},
year = {2024},
publisher = {JMLR.org},
booktitle = {ICML},
articleno = {297},
numpages = {19},
location = {Vienna, Austria},
series = {ICML'24}
}

@inproceedings{effort,
  title={Effort: Efficient Orthogonal Modeling for Generalizable AI-Generated Image Detection},
  author={Yan, Zhiyuan and Wang, Jiangming and Wang, Zhendong and Jin, Peng and Zhang, Ke-Yue and Chen, Shen and Yao, Taiping and Ding, Shouhong and Wu, Baoyuan and Yuan, Li},
  booktitle = {ICML},
  year={2024}
}

@inproceedings{jiang2025tabdsr,
  title={TABDSR: Decompose, Sanitize, and Reason for Complex Numerical Reasoning in Tabular Data},
  author={Jiang, Changjiang and Yu, Fengchang and Chen, Haihua and Lu, Wei and Zeng, Jin},
  booktitle={Findings of the Association for Computational Linguistics: EMNLP 2025},
  pages={3172--3196},
  year={2025}
}

@inproceedings{fakehr1,
      title={Fake-HR1: Rethinking Reasoning of Vision Language Model for Synthetic Image Detection}, 
      author={Changjiang Jiang and Xinkuan Sha and Fengchang Yu and Jingjing Liu and Jian Liu and Mingqi Fang and Chenfeng Zhang and Wei Lu},
      year={2026},
      booktitle={ICASSP},
      url={https://arxiv.org/abs/2602.10042} 
}

@article{ying2026beyond,
    title={Beyond Human Annotation: Recent Advances in Data Generation Methods for Document Intelligence},
    author={Ying, Dehao and Yu, Fengchang and Chen, Haihua and Jiang, Changjiang and Li, Yurong and Lu, Wei},
    journal={arXiv preprint arXiv:2601.12318},
    year={2026}
}

@article{xin2026hytrechybridtemporalawareattention,
      title={HyTRec: A Hybrid Temporal-Aware Attention Architecture for Long Behavior Sequential Recommendation}, 
      author={Lei Xin and Yuhao Zheng and Ke Cheng and Changjiang Jiang and Zifan Zhang and Fanhu Zeng},
      year={2026},
      journal={arXiv preprint arXiv:2602.18283},
      url={https://arxiv.org/abs/2602.18283}, 
}

@inproceedings{
    zhao2026spinebench,
    title={SpineBench: A Clinically Salient, Level-Aware Benchmark Powered by the SpineMed-450k Corpus},
    author={Ming Zhao and Wenhui Dong and Yang Zhang and XiangZheng and Zhonghao Zhang and Zian Zhou and YUNZHI GUAN and Liukun Xu and Wei Peng and Zhaoyang Gong and Zhicheng Zhang and Dachuan li and Xiaosheng Ma and Yuli Ma and Jianing Ni and Changjiang Jiang and Lixia Tian and Chen Qixin and Xia Kaishun and Pingping Liu and Tongshun Zhang and ZhiqiangLiu and Zhongan Bi and Chenyang Si and Tiansheng Sun and Caifeng Shan},
    booktitle={The Fourteenth International Conference on Learning Representations},
    year={2026},
    url={https://arxiv.org/abs/2601.12318}
}

@article{du2025forensichub,
  title={Forensichub: A unified benchmark \& codebase for all-domain fake image detection and localization},
  author={Du, Bo and Zhu, Xuekang and Ma, Xiaochen and Qu, Chenfan and Feng, Kaiwen and Yang, Zhe and Pun, Chi-Man and Liu, Jian and Zhou, Ji-Zhe},
  journal={arXiv preprint arXiv:2505.11003},
  year={2025}
}

@inproceedings{yuanse,
  title={SE-GUI: Enhancing Visual Grounding for GUI Agents via Self-Evolutionary Reinforcement Learning},
  author={Yuan, Xinbin and Zhang, Jian and Li, Kaixin and Cai, Zhuoxuan and Yao, Lujian and Chen, Jie and Wang, Enguang and Hou, Qibin and Chen, Jinwei and Jiang, Peng-Tao and others},
  booktitle={The Thirty-ninth Annual Conference on Neural Information Processing Systems},
  year={2025}
}

@inproceedings{lin2025audio,
  title={Audio Does Matter: Importance-Aware Multi-Granularity Fusion for Video Moment Retrieval},
  author={Lin, Junan and Liu, Daizong and Chen, Xianke and Qu, Xiaoye and Yang, Xun and Zhu, Jixiang and Zhang, Sanyuan and Dong, Jianfeng},
  booktitle={Proceedings of the 33rd ACM International Conference on Multimedia},
  pages={6027--6036},
  year={2025}
}

@inproceedings{qu2023doctamper,
    author    = {Qu, Chenfan and Liu, Chongyu and Liu, Yuliang and Chen, Xinhong and Peng, Dezhi and Guo, Fengjun and Jin, Lianwen},
    title     = {Towards Robust Tampered Text Detection in Document Image: New Dataset and New Solution},
    booktitle = {Proceedings of the IEEE/CVF Conference on Computer Vision and Pattern Recognition (CVPR)},
    month     = {June},
    year      = {2023},
    pages     = {5937-5946}
}

@inproceedings{qu2024miml,
    author    = {Qu, Chenfan and Zhong, Yiwu and Liu, Chongyu and Xu, Guitao and Peng, Dezhi and Guo, Fengjun and Jin, Lianwen},
    title     = {Towards Modern Image Manipulation Localization: A Large-Scale Dataset and Novel Methods},
    booktitle = {Proceedings of the IEEE/CVF Conference on Computer Vision and Pattern Recognition (CVPR)},
    month     = {June},
    year      = {2024},
    pages     = {10781-10790}
}

@inproceedings{qu2025ostf,
  title={Revisiting tampered scene text detection in the era of generative AI},
  author={Qu, Chenfan and Zhong, Yiwu and Guo, Fengjun and Jin, Lianwen},
  booktitle={Proceedings of the AAAI Conference on Artificial Intelligence},
  volume={39},
  pages={694-702},
  year={2025}
}

@inproceedings{
qu2024omni,
title={Omni-{IML}: Towards Unified Interpretable Image Manipulation Localization},
author={Chenfan Qu and Yiwu Zhong and Fengjun Guo and Lianwen Jin},
booktitle={The Fourteenth International Conference on Learning Representations},
year={2026},
}

@inproceedings{qu2026textshield,
  title={TextShield-R1: Reinforced Reasoning for Tampered Text Detection},
  author={Qu, Chenfan and Zhong, Yiwu and Liu, Jian and Zhu, Xuekang and Yu, Bohan and Jin, Lianwen},
  booktitle={Proceedings of the AAAI Conference on Artificial Intelligence},
  volume={40},
  pages={8621-8629},
  year={2026}
}

@inproceedings{zhu2025mesorch,
  title={Mesoscopic insights: Orchestrating multi-scale \& hybrid architecture for image manipulation localization},
  author={Zhu, Xuekang and Ma, Xiaochen and Su, Lei and Jiang, Zhuohang and Du, Bo and Wang, Xiwen and Lei, Zeyu and Feng, Wentao and Pun, Chi-Man and Zhou, Ji-Zhe},
  booktitle={Proceedings of the AAAI Conference on Artificial Intelligence},
  volume={39},
  number={10},
  pages={11022--11030},
  year={2025}
}

@article{ma2025imdl,
  title={Imdl-benco: A comprehensive benchmark and codebase for image manipulation detection \& localization},
  author={Ma, Xiaochen and Zhu, Xuekang and Su, Lei and Du, Bo and Jiang, Zhuohang and Tong, Bingkui and Lei, Zeyu and Yang, Xinyu and Pun, Chi-Man and Lv, Jiancheng and others},
  journal={Advances in Neural Information Processing Systems},
  volume={37},
  pages={134591--134613},
  year={2025}
}

@article{zhu2025does,
  title={Does the Manipulation Process Matter? RITA: Reasoning Composite Image Manipulations via Reversely-Ordered Incremental-Transition Autoregression},
  author={Zhu, Xuekang and Zhou, Ji-Zhe and Feng, Kaiwen and Qu, Chenfan and Wang, Yunfei and Zhou, Liting and Liu, Jian},
  journal={arXiv preprint arXiv:2509.20006},
  year={2025}
}

@article{kong2025token,
  title={Token Reduction Should Go Beyond Efficiency in Generative Models--From Vision, Language to Multimodality},
  author={Kong, Zhenglun and Li, Yize and Zeng, Fanhu and Xin, Lei and Messica, Shvat and Lin, Xue and Zhao, Pu and Kellis, Manolis and Tang, Hao and Zitnik, Marinka},
  journal={arXiv preprint arXiv:2505.18227},
  year={2025}
}

@article{xin2024artificial,
  title={Artificial intelligence for central dogma-centric multi-omics: Challenges and breakthroughs},
  author={Xin, Lei and Huang, Caiyun and Li, Hao and Huang, Shihong and Feng, Yuling and Kong, Zhenglun and Liu, Zicheng and Li, Siyuan and Yu, Chang and Shen, Fei and others},
  journal={arXiv preprint arXiv:2412.12668},
  year={2024}
}

@inproceedings{xin2026dualcpt,
  title={DualCPT: Dual-branch Modeling for Cellular Phenotype Transition},
  author={Xin, Lei and Kong, Zhenglun and Chen, Fukang and Wang, Yuhao Zheng4 Zeheng and Tang, Hao},
  booktitle={AAAI Bridge Program on AI for Medicine and Healthcare},
  pages={302--312},
  year={2026},
  organization={PMLR}
}

@article{zhao2026know,
  title={Know What You Know: Metacognitive Entropy Calibration for Verifiable RL Reasoning},
  author={Zhao, Qiannian and Yang, Chen and Jing, Jinhao and Zhang, Yunke and Ren, Xuhui and Yu, Lu and Zhang, Shijie and Yin, Hongzhi},
  journal={arXiv preprint arXiv:2602.22751},
  year={2026}
}

@article{huang2025unishield,
  title={UniShield: An Adaptive Multi-Agent Framework for Unified Forgery Image Detection and Localization},
  author={Huang, Qing and Xu, Zhipei and Zhang, Xuanyu and Zhang, Jian},
  journal={arXiv preprint arXiv:2510.03161},
  year={2025}
}

@article{yu2026agentfox,
  title={AgentFoX: LLM Agent-Guided Fusion with eXplainability for AI-Generated Image Detection},
  author={Yu, Yangxin and Zhou, Yue and Li, Bin and others},
  journal={arXiv preprint arXiv:2603.23115},
  year={2026}
}

@article{Zhou2025AIGIHolmesTE,
  title={AIGI-Holmes: Towards Explainable and Generalizable AI-Generated Image Detection via Multimodal Large Language Models},
  author={Ziyin Zhou and Yunpeng Luo and Yuanchen Wu and Ke Sun and Jiayi Ji and Ke Yan and Shouhong Ding and Xiaoshuai Sun and Yunsheng Wu and Rongrong Ji},
  journal={ArXiv},
  year={2025},
  volume={abs/2507.02664},
  url={https://api.semanticscholar.org/CorpusID:280141523}
}

@inproceedings{tan2025veritas,
  title={Veritas: Generalizable Deepfake Detection via Pattern-Aware Reasoning},
  author={Tan, Hao and Lan, Jun and Tan, Zichang and Liu, Ajian and Song, Chuanbiao and Shi, Senyuan and Zhu, Huijia and Wang, Weiqiang and Wan, Jun and Lei, Zhen},
  booktitle={International Conference on Learning Representations},
  year={2026}
}

@article{tan2026videoveritas,
 	title={VideoVeritas: AI-Generated Video Detection via Perception Pretext Reinforcement Learning},
  author={Tan, Hao and Lan, Jun and Shi, Senyuan and Tan, Zichang and Yu, Zijian and Zhu, Huijia and Wang, Weiqiang and Wan, Jun and Lei, Zhen},
  journal={arXiv preprint arXiv:2602.08828},
  year={2026}
}

@article{liu2026mirror,
  title={MIRROR: Manifold Ideal Reference ReconstructOR for Generalizable AI-Generated Image Detection},
  author={Liu, Ruiqi and Cui, Manni and Qin, Ziheng and Yan, Zhiyuan and Chen, Ruoxin and Han, Yi and Li, Zhiheng and Chen, Junkai and Chen, ZhiJin and Lin, Kaiqing and others},
  journal={arXiv preprint arXiv:2602.02222},
  year={2026}
}

@article{zhou2025breaking,
  title={Breaking latent prior bias in detectors for generalizable aigc image detection},
  author={Zhou, Yue and He, Xinan and Lin, KaiQing and Fan, Bin and Ding, Feng and Li, Bin},
  journal={arXiv preprint arXiv:2506.00874},
  year={2025}
}

@article{jiang2025revisiting,
  title={Revisiting Reconstruction-based AI-generated Image Detection: A Geometric Perspective},
  author={Jiang, Wan and Yan, Jing and Zhang, Ruixuan and Chen, Xiaojing and Miao, Changtao and Li, Zhe and Lin, Chenhao and Diao, Yunfeng and Hong, Richang},
  journal={arXiv e-prints},
  pages={arXiv--2510},
  year={2025}
}

@article{li2025artificial,
  title={Is artificial intelligence generated image detection a solved problem?},
  author={Li, Ziqiang and Yan, Jiazhen and He, Ziwen and Zeng, Kai and Jiang, Weiwei and Xiong, Lizhi and Fu, Zhangjie},
  journal={arXiv preprint arXiv:2505.12335},
  year={2025}
}

@article{ji2025zoom,
  title={Zoom-In to Sort AI-Generated Images Out},
  author={Ji, Yikun and Hong, Yan and Deng, Bowen and Zhu, Huijia and Wang, Weiqiang and Zhang, Liqing and Zhang, Jianfu and others},
  journal={arXiv preprint arXiv:2510.04225},
  year={2025}
}

@article{zhang2025detecting,
  title={Detecting Generated Images by Fitting Natural Image Distributions},
  author={Zhang, Yonggang and Nie, Jun and Tian, Xinmei and Gong, Mingming and Zhang, Kun and Han, Bo},
  journal={arXiv preprint arXiv:2511.01293},
  year={2025}
}

@article{shuai2026detectors,
  title={When Detectors Forget Forensics: Blocking Semantic Shortcuts for Generalizable AI-Generated Image Detection},
  author={Shuai, Chao and Liu, Zhenguang and Fan, Shaojing and Gong, Bin and Lian, Weichen and Bi, Xiuli and Ba, Zhongjie and Ren, Kui},
  journal={arXiv preprint arXiv:2603.09242},
  year={2026}
}

@article{park2025vidguard,
  title={Vidguard-r1: Ai-generated video detection and explanation via reasoning mllms and rl},
  author={Park, Kyoungjun and Yang, Yifan and Yi, Juheon and Zheng, Shicheng and Shen, Yifei and Han, Dongqi and Shan, Caihua and Muaz, Muhammad and Qiu, Lili},
  journal={arXiv preprint arXiv:2510.02282},
  year={2025}
}

@inproceedings{ying2024dig,
  title={Dig: Complex layout document image generation with authentic-looking text for enhancing layout analysis},
  author={Ying, Dehao and Yu, Fengchang and Chen, Haihua and Lu, Wei},
  booktitle={Proceedings of the 32nd ACM International Conference on Multimedia},
  pages={3239--3247},
  year={2024}
}

@inproceedings{ying2024fine,
  title={Fine-Grained, Accurate Data Generation and Multimodal Layout Analysis for Academic Papers},
  author={Ying, Dehao and Yu, Fengchang and Chen, Haihua and Lu, Wei},
  booktitle={Proceedings of the 24th ACM/IEEE Joint Conference on Digital Libraries},
  pages={1--11},
  year={2024}
}

@article{wang2026forgeryvcr,
  title={ForgeryVCR: Visual-Centric Reasoning via Efficient Forensic Tools in MLLMs for Image Forgery Detection and Localization},
  author={Wang, Youqi and Chen, Shen and Wang, Haowei and Peng, Rongxuan and Yao, Taiping and Tan, Shunquan and Chen, Changsheng and Li, Bin and Ding, Shouhong},
  journal={arXiv preprint arXiv:2602.14098},
  year={2026}
}

@article{zhoubao-earlvr,
      title={Incentivizing Parametric Knowledge via Reinforcement Learning with Verifiable Rewards for Cross-Cultural Entity Translation}, 
      author={Jiang Zhou and Xiaohu Zhao and Xinwei Wu and Tianyu Dong and Hao Wang and Yangyang Liu and Heng Liu and Linlong Xu and Longyue Wang and Weihua Luo and Deyi Xiong},
      year={2026},
      journal={arXiv preprint arXiv:2604.16881},
}

\newpage
\beginappendix

\section{Author and Affiliation List}
\label{sec:contri}

\noindent \textbf{Nanjing University} \quad 
Wenhui Dong\textsuperscript{*}\textsuperscript{\Letter},
Chenyang Si, 
Caifeng Shan

\noindent \textbf{Wuhan University} \quad 
Changjiang Jiang\textsuperscript{*}\textsuperscript{+}, 
Fengchang Yu

\noindent \textbf{Stanford University} \quad 
Wei Peng

\noindent \textbf{Ningxia University} \quad 
Zhonghao Zhang

\noindent \textbf{Nankai University} \quad 
Xinbin Yuan

\noindent \textbf{Georgia Institute of Technology} \quad 
Yifei Bi

\noindent \textbf{Jilin University} \quad 
Ming Zhao

\noindent \textbf{Zhejiang University} \quad 
Zian Zhou

\noindent \textsuperscript{*} Equal contribution. \textsuperscript{\Letter} Corresponding author. \\
\textsuperscript{+} This work was done while serving as a research intern at Nanjing University.\\

\section{License}
\label{sec:license}

For all datasets in our experiments and datasets, FLUX.1~\citep{labs2025flux1kontextflowmatching} and Lora Flux~\citep{labs2025flux1kontextflowmatching} falls under the FLUX.1 [dev] Non-Commercial License.WildFake~\citep{wildfake_2025} is released under an open-source license, making it freely available for research and non-commercial use; Qwen-Image~\citep{wu2025qwenimagetechnicalreport} is licensed under Apache 2.0; Sora~\citep{sora_videoworldsimulators2024} is a commercial, non-open-source text-to-video model provided via ChatGPT Plus/Pro; user-generated content is owned by the user and allows (subject to usage policies) non-commercial use, but the model itself is closed-source. 

Stable Diffusion~\citep{svd_blattmann2023stable}, while originally open-source under Creative ML OpenRAIL-M (e.g., versions 1.X, 2.1, SDXL), has transitioned in newer versions (SD3.x, SD3.5) to the Stability AI Community License, which permits free use for individuals or entities with annual revenue under USD 1 million, but requires a paid Enterprise License for larger-scale or commercial usage.

During \benchmark~construction and experimentation, we accessed Gemini API and ChatGPT API to ensure annotation and evaluation's quality and reproducibility. We explicitly state that all data generated using these commercial APIs are used solely for academic research and non-commercial purposes, fully complying with the respective API usage agreements and ethical guidelines. No Gemini or ChatGPT derived content is redistributed for any commercial training, deployment, or monetization purposes.

\section{Related Work}
\label{appendix:relatedwork}

\subsection{Methods for Synthetic Content Detection}
Due to growing concerns about the misuse of synthetic data~\citep{deng2024survey}, research on AI-generated content (AIGC) detection has expanded rapidly in recent years. Most existing models for AI-generated images and videos formulate the task as binary classification, simply predicting whether the content is "real" or "fake." Representative examples include CNN-based AIGVDet~\citep{aigvdet_bai2024ai}, CNNSpot~\citep{cnnspot_wang2020cnn} and Transformer-based models such as DIRE~\citep{dire_wang2023dire} and AIDE~\citep{aide_yan2025a}. Meanwhile, several works have explored the application of multimodal large language models (MLLMs) to AIGC detection, including Synartifact~\citep{cao2024synartifact} and Bi-LORA~\citep{keita2025bi}. However, these approaches largely overlook the importance of interpretability in AIGC detection.

Some efforts attempt to introduce interpretability by leveraging spatial annotations~\citep{dong2022explaining} or frequency-domain artifact analysis~\citep{zhang2023perceptual}. Nevertheless, the resulting explanations are often difficult for humans to comprehend, as they lack clarity in natural language. This limitation is particularly evident in the video domain, where AI-generated content frequently exhibits obvious flaws, e.g., incoherent frame transitions and object inconsistency, that are easily noticed and reasoned about by humans~\citep{deng2024survey}. FakeClue~\citep{fakeclue_wen2025spot} introduces the use of vision-language models (VLMs) to provide interpretability for image-level detection, but it does not offer a unified framework that integrates both images and videos.

\begin{figure}[t]
\begin{tcolorbox}[
  colback=white, 
  colframe=black, 
  colupper=gray!70, 
  fontupper=\footnotesize, 
  coltitle=white, 
  coltext=black, 
  boxrule=0.5mm, 
  title=Image Authenticity Analyst Assistant (User Prompt)
]

Is this image real or fake?

\end{tcolorbox}
\caption{User Prompt Template for Image Data Distillation}
\label{prompt:data_distil_user_1}
\end{figure}

\begin{figure}[ht]
\begin{tcolorbox}[
  colback=white, 
  colframe=black, 
  colupper=gray!70, 
  fontupper=\footnotesize, 
  coltitle=white, 
  coltext=black, 
  boxrule=0.5mm, 
  title=Video Authenticity Analyst Assistant (User Prompt)
]

Is this video real or fake?

\end{tcolorbox}
\caption{User Prompt Template for Video Data Distillation}
\label{prompt:data_distil_user_video_2}
\end{figure}

\begin{figure*}[htbp]
\begin{tcolorbox}[
  colback=white, 
  colframe=black, 
  colupper=gray!70, 
  fontupper=\footnotesize, 
  coltitle=white, 
  coltext=black, 
  boxrule=0.5mm, 
  title=Image Authenticity Analyst Assistant (System Prompt)
]
\#\# Role\\
Expert AI system for detecting image by analyzing visual anomalies across \textbf{spatial plausibility}.\\

\#\# Analysis Dimensions\\
\#\#\# Spatial Features: static anomaly detection\\
- \textbf{Impractical Luminosity}\\
\hspace*{1em}- Scene brightness measurement\\
\hspace*{1em}- Invisible light source detection (physical validation) \\ 
- \textbf{Localized Blur}\\
\hspace*{1em}- Focus distribution mapping (sharpness gradient)  \\
\hspace*{1em}- Artificial depth-of-field identification (algorithmic artifacts)  \\
- \textbf{Illegible Letters}\\
\hspace*{1em}- OCR text extraction\\
\hspace*{1em}- Character structural integrity (stroke continuity) \\ 
- \textbf{Distorted Components}\\
\hspace*{1em}- Anatomical/proportional accuracy (biological/object logic)  \\
\hspace*{1em}- Physics compliance (material/gravity validation)  \\
- \textbf{Omitted Components}\\
\hspace*{1em}- Object completeness check (edge/detail absence)  \\
\hspace*{1em}- Partial rendering artifact detection (AI-generated traces)  \\
- \textbf{Spatial Relationships}\\
\hspace*{1em}- Contextual object placement (scene plausibility)  \\
\hspace*{1em}- Perspective consistency (geometric projection)  \\
- \textbf{Chromatic Irregularity}\\
\hspace*{1em}- Color database comparison (natural distribution)\\
\hspace*{1em}- Unnatural hue detection (oversaturation/abrupt gradients)\\
- \textbf{Abnormal Texture}\\
\hspace*{1em}- Surface pattern regularity (texture repetition)\\
\hspace*{1em}- Material property coherence (reflectance/roughness validation)\\

\#\# Reasoning Step\\
1. \textbf{Spatial Analysis}\\
\hspace*{1em}- Analyze static features (e.g., lighting, text, objects)  \\
2. \textbf{Conclusion}: Only real or fake.\\
\hspace*{1em}- real: Contains verifiable capture device signatures and natural physical imperfections.\\
\hspace*{1em}- fake: Exhibits synthetic fingerprints including but not limited to over-regularized textures and non-physical light interactions.\\

The assistant first thinks about the reasoning step in the mind and then provides the user with the reason. The reasoning step and conclusion are enclosed within \verb|<think>| \verb|</think>| and \verb|<conclusion>| \verb|</conclusion>| tags, respectively, i.e., \verb|<think>| reasoning step here \verb|</think>| \verb|<conclusion>| real or fake \verb|</conclusion>|. \verb|<conclusion>| content must strictly align with the user-provided authenticity label (real/fake) in both value and semantic context.

\end{tcolorbox}
\caption{System Prompt Template for Image Data Distillation}
\label{prompt:data_distil_system_1}
\end{figure*}

\begin{figure*}[ht]
  \centering
\adjustbox{max height=0.9\textheight}{%
\begin{tcolorbox}[
  colback=white, 
  colframe=black, 
  colupper=gray!70, 
  fontupper=\footnotesize, 
  coltitle=white, 
  coltext=black, 
  boxrule=0.5mm, 
  title=Video Authenticity Analyst Assistant (System Prompt)
]
\#\# Role\\
Expert AI system for detecting videos by analyzing visual anomalies across \textbf{temporal coherence} (inter-frame dynamics) and \textbf{spatial plausibility} (intra-frame logic).\\

\#\# Analysis Dimensions\\
\#\#\# 1. Temporal Features: Multi-frame dynamic anomaly detection
- \textbf{Luminance Discrepancy}\\
\hspace*{1em}- Shadow direction consistency (cross-frame comparison)\\
\hspace*{1em}- Light source coordination (temporal validation)\\
- \textbf{Awkward Facial Expression}\\
\hspace*{1em}- Facial muscle motion continuity (expression dynamics)\\
\hspace*{1em}- Emotion-context alignment (temporal coherence)\\
- \textbf{Duplicated Components}\\
\hspace*{1em}- Repeating element pattern recognition (cross-frame tracking)\\
\hspace*{1em}- Natural variation analysis (sequence validation)\\
- \textbf{Non-Spatial Relationships}\\
\hspace*{1em}- Object interaction physics (motion trajectory validation)\\
\hspace*{1em}- Fusion/penetration anomalies (temporal detection)\\

\#\#\# 2. Spatial Features: Single-frame static anomaly detection\\
- \textbf{Impractical Luminosity}\\
\hspace*{1em}- Scene brightness measurement (single-frame analysis)  \\
\hspace*{1em}- Invisible light source detection (physical validation)  \\
- \textbf{Localized Blur}\\
\hspace*{1em}- Focus distribution mapping (sharpness gradient)  \\
\hspace*{1em}- Artificial depth-of-field identification (algorithmic artifacts) \\ 
- \textbf{Illegible Letters}\\
\hspace*{1em}- OCR text extraction (single-frame recognition) \\ 
\hspace*{1em}- Character structural integrity (stroke continuity)\\  
- \textbf{Distorted Components}\\
\hspace*{1em}- Anatomical/proportional accuracy (biological/object logic) \\ 
\hspace*{1em}- Physics compliance (material/gravity validation)  \\
- \textbf{Omitted Components}\\
\hspace*{1em}- Object completeness check (edge/detail absence)  \\
\hspace*{1em}- Partial rendering artifact detection (AI-generated traces)\\  
- \textbf{Spatial Relationships}\\
\hspace*{1em}- Contextual object placement (scene plausibility)  \\
\hspace*{1em}- Perspective consistency (geometric projection)  \\
- \textbf{Chromatic Irregularity}\\
\hspace*{1em}- Color database comparison (natural distribution)  \\
\hspace*{1em}- Unnatural hue detection (oversaturation/abrupt gradients) \\ 
- \textbf{Abnormal Texture}\\
\hspace*{1em}- Surface pattern regularity (texture repetition)  \\
\hspace*{1em}- Material property coherence (reflectance/roughness validation)\\

\#\# Reasoning Step\\
1. \textbf{Temporal Analysis}\\
\hspace*{1em}- Track dynamic features across frames (e.g., shadows, expressions)\\
2. \textbf{Spatial Analysis}\\
\hspace*{1em}- Analyze static features per frame (e.g., lighting, text, objects)  \\
3. \textbf{Conclusion}: Only real or fake.\\
\hspace*{1em}- real: Contains verifiable capture device signatures and natural physical imperfections.\\
\hspace*{1em}- fake: Exhibits synthetic fingerprints including but not limited to over-regularized textures and non-physical light interactions.\\

The assistant first thinks about the reasoning step in the mind and then provides the user with the reason. The reasoning step and conclusion are enclosed within \verb|<think>| \verb|</think>| and \verb|<conclusion>| \verb|</conclusion>| tags, respectively, i.e., \verb|<think>| reasoning step here \verb|</think>| \verb|<conclusion>| real or fake \verb|</conclusion>|. \verb|<conclusion>| content must strictly align with the user-provided authenticity label (real/fake) in both value and semantic context.

\end{tcolorbox}%
  }
\caption{System Prompt Template for Video Data Distillation}
\label{prompt:data_distil_system_video_1}
\end{figure*}

\subsection{Datasets for Synthetic Content Detection.}

Early datasets for synthetic content detection, such as CNNSpot~\citep{cnnspot_wang2020cnn}, primarily collected fake images generated by GAN-based models~\citep{gan1_goodfellow2014generative,gan2_zhu2017unpaired,gan3_brock2018large}. However, with the advent of more advanced generative architectures like diffusion models~\citep{diff1_ho2020denoising,diff2_dhariwal2021diffusion,diff5_nichol2021glide} and their variants, the authenticity of generated content has significantly increased, making it more challenging for detection models to discern. This has spurred the development of newer datasets, including ArtiFact~\citep{cao2024synartifact}, GenImage~\citep{genimage_2023}, and WildFake~\citep{wildfake_2025}. GenImage~\citep{genimage_2023} comprises images from the 1000 ImageNet~\citep{imagenet_ILSVRC15} categories, generated by eight state-of-the-art generators. Nevertheless, these datasets predominantly focus on image-based content. More recently, datasets emphasizing interpretability have also been introduced. FakeClue~\citep{fakeclue_wen2025spot} contains a large amount of image data with explainability annotations but lacks video data. LOKI~\citep{loki_ye2024loki} offers data across 26 different categories and includes 18,000 distinct questions; however, its volume of multimodal data is relatively small and primarily suited for evaluation rather than comprehensive model training. Therefore, a critical gap exists for a unified benchmark encompassing both image and video modalities to rigorously evaluate the performance of contemporary AIGC detectors.

\begin{figure*}[ht]
    \centering
    \includegraphics[width=0.95\textwidth]{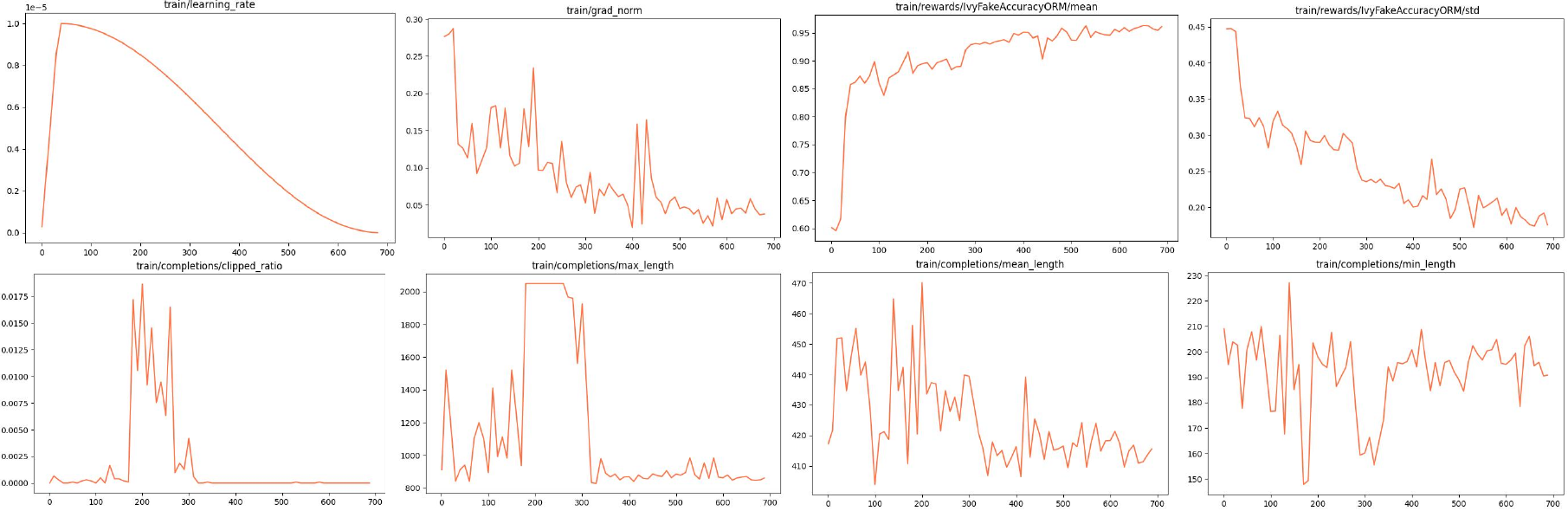}
    \caption{Final Reinforcement Learning Training Curves.}
    \label{fig:training}
\end{figure*}

\section{GRPO}
\label{appendix:grpo}

Following DeepSeek-R1~\citep{deepseekr1_guo2025deepseek}, we adapt the Group Relative Policy Optimization (GRPO)~\citep{shao2024deepseekmathpushinglimitsmathematical}, an online RL algorithm designed to maximize the advantage of generated completions while constraining policy divergence from a reference model. We formalize our training process of \baseline using GRPO below. Let $p$ denote a sampled prompt, and let ${o_{1}, o_{2}, . . . , o_{n}}$ be a group of completions generated by the current policy $\pi_{\theta}$. For each completion $G_{i}$, a reward $r_{i}$ is computed using a rule-based reward function. The group advantage for each completion is then calculated as:

\begin{equation}
\hat{A}_{i,t}=\frac{r_{i}-mean(r)}{std(r)}
\end{equation}

\begin{figure*}[t]
    \begin{equation}
\mathcal{L}_{\mathrm{GRPO}}(\theta)
= -\frac{1}{K}\sum_{k=1}^{K} A^{(k)}\,\ell^{(k)}(\theta)
\;+\; \beta\,\mathrm{KL}\!\left(\pi_{\theta}(\cdot|x)\,\|\,\pi_{\mathrm{ref}}(\cdot|x)\right).
\end{equation}
\end{figure*}


where $\beta$ is a coefficient that balances advantage maximization and KL regularization, and the clipping operator $clip(\dots, 1-\epsilon, 1+\epsilon)$ constrains the update magnitude. To regularize policy updates, we estimate the token-level Kullback-Leibler (KL) divergence between the current policy $\pi_{\theta}$ and a reference policy $\pi_{ref}$.

\begin{figure*}[t]
\begin{equation}
\mathbb{D}_{\text{KL}} \big[ \pi_\theta \big\| \pi_{\text{ref}} \big] = \frac{\pi_{\text{ref}}(o_{i,t} \mid p, o_{i,<t})}{\pi_\theta(o_{i,t} \mid p, o_{i,<t})} - \log \frac{\pi_{\text{ref}}(o_{i,t} \mid p, o_{i,<t})}{\pi_\theta(o_{i,t} \mid p, o_{i,<t})} - 1.
\label{fig:eq_kl}
\end{equation}
\end{figure*}

\textbf{Reward Model}. For effective RL, we employ a rule-based reward that consists of accuracy and format rewards. We introduce a solid accuracy reward
for AIGC Detection, which utilizes distinct functions to evaluate binary classification task. This allows for a more appropriate assessment based on the expected answer type.

\begin{itemize}
    \item \textbf{Accuracy Reward}: The accuracy reward assigns a score of \textbf{1} if the label in \texttt{<conclusion>} exactly matches the ground-truth classification \texttt{real} and \texttt{fake} and \textbf{0} otherwise.
    \item \textbf{Format Reward}: The format reward assigns a score of \textbf{1} if the output strictly follows the structural requirements by enclosing the reasoning within \texttt{<think></think>} tags and the final decision within \texttt{<conclusion></conclusion>} tags, and \textbf{0} otherwise.
\end{itemize}

\section{Effect of Incorporating Human-Annotated Labels via \texttt{gemini 2.5 pro} on Accuracy}
\label{appendix:human_eva}

To assess the impact of human-annotated labels on model performance, we compare the accuracy of final conclusion predictions under two settings: (i) with labels incorporated via the \texttt{gemini 2.5 pro}, and (ii) without labels. The evaluation was conducted on about 1,000 examples from the test set.

\begin{table}[!h]
\centering
\begin{tabular}{l c}
\toprule
\textbf{Annotation Setting} & \textbf{Accuracy (Acc)} \\
\midrule
With Label & 1.000 \\
Without Label & 0.785 \\
\bottomrule
\end{tabular}
\caption{Accuracy of conclusion prediction with and without incorporating labels.}
\label{tab:label}
\end{table}

As shown in Table~\ref{tab:label}, incorporating ground-truth labels results in a substantial performance gain, yielding perfect accuracy (1.000), compared to 0.785 without labels. The drastic performance gap suggests potential limitations in label-free or weakly supervised setups when applied to tasks requiring fine-grained semantic understanding.

\section{Data Distribution}
\label{appendix:dd}


\section{Additional Experiment}
\label{appendix:ae}

\subsection{Synthetic Image Detection}

We evaluated our AIGC detector on the GenImage~\citep{genimage_2023} and Chameleon~\citep{aide_yan2025a} benchmarks. 

As shown in Table~\ref{tab:genimage}, GenImage comprises seven subsets generated by leading models, i.e., Midjourney, Stable Diffusion v1.4 \& v1.5, ADM, GLIDE, Wukong, VQDM, and BigGAN. We compared against five state-of-the-art detectors, which are CNNSpot~\citep{cnnspot_wang2020cnn}, F3Net~\citep{qian2020thinking}, DIRE~\citep{dire_wang2023dire}, GenDet~\citep{zhu2023gendet}, PatchCraft~\citep{aigcdetectionbenchmark_zhong2023patchcraft}, and AIDE~\citep{aide_yan2025a}. 

Notably, our detector surpasses previous state-of-the-art methods such as AIDE (86.88\%) and PatchCraft (82.30\%) by a large margin, even though it is trained under a unified multimodal paradigm rather than a generator-specific setting. The improvement is especially pronounced on challenging generators like \textit{ADM}, \textit{GLIDE}, and \textit{BigGAN}, where conventional CNN- or patch-based detectors often fail to capture high-frequency inconsistencies or generalized texture patterns. 

\textbf{Notably, our model is trained with only 130K image samples—fewer than GenImage’s 2M+ training images—yet still achieves superior generalization performance.}


\subsection{Synthetic Video Detection}

As shown in Table~\ref{tab:genvideo}, we evaluate the performance of our detector on the GenVideo benchmark~\citep{demamba}.

Specifically, \baseline~achieves an average F1 score of \textbf{95.26\%} and recall of \textbf{95.28\%}, substantially surpassing the prior state-of-the-art DeMamba (F1 = 90.20\%). The improvement is particularly evident on challenging categories such as \textit{HotShot}, \textit{Lavie}, and \textit{Show-1}, where previous methods tend to overfit to specific generative distributions or temporal artifacts. Our results indicate that the unified MLLM-based detection paradigm not only captures spatial inconsistencies but also learns transferable temporal-spatial representations, leading to more robust generalization across unseen generative models.

\textbf{Notably, our model is trained with only 64K video samples—over 35$\times$ fewer than DeMamba’s 2.295M training videos—yet still achieves superior generalization performance.}

\section{Prompts}
\label{appendix:prompt}

\begin{figure*}[ht]
\begin{tcolorbox}[
  colback=white, 
  colframe=black, 
  colupper=gray!70, 
  fontupper=\footnotesize, 
  coltitle=white, 
  coltext=black, 
  boxrule=0.5mm, 
  title=Default System Prompt
]

You are an AI-generated content detector. Given a single media (image or video), classify it as real or fake. Provide detailed reasoning inside the \verb|<think>|...\verb|</think>| tags, including your step-by-step thought process.\\

Your output must begin with \verb|<think>|\\n and end with \verb|</conclusion>|.\\

Then output exactly one word in lowercase—either real or fake—wrapped in \verb|<conclusion>|...\verb|</conclusion>|.\\

Do not include any other words. If uncertain, choose the most likely class.

\end{tcolorbox}
\caption{Default System Prompt for Training and Evaluation}
\label{prompt:system}
\end{figure*}

Here we provide the prompts that are mainly used in this study. The default system prompt can be seen in Figure\ref{prompt:system}. As illustrated by the following figures, there are five distillation prompts distillation we used in this paper that mainly can be divided into the following three parts:

\textbf{Prompt Template for Image Data Distillation}: Since image data consists of a single frame, it can be treated as a static instance. Therefore, AIGC detection mainly focuses on identifying spatial anomalies. Detail prompt can be found in Figure~\ref{prompt:data_distil_user_1} and ~\ref{prompt:data_distil_system_1}.

\textbf{Prompt Template for Video Data Distillation}: Compared to images, video inputs provide continuous multi-frame context. This allows for detection along both spatial and temporal anomaly dimensions. Dtail prompt can be found in Figure~\ref{prompt:data_distil_user_video_2} and Figure~\ref{prompt:data_distil_system_video_1}.

\textbf{GPT Assisted Evaluation Prompt}: To assess the quality of model outputs, we design a GPT-based evaluator prompt that scores responses across four dimensions: Completeness, Relevance, Level of Detail, and Explanation. The evaluator receives a structured pair of GroundTruth and ModelOutput, each containing a \verb|<think>| section (reasoning) and a \verb|<conclusion>| (final judgment). The model must return a structured JSON object with integer scores (1–5) for each dimension. The prompt is provided in Figure~\ref{prompt:gpt_eva}.

\begin{figure*}[ht]
\begin{tcolorbox}[
  colback=white, 
  colframe=black, 
  colupper=gray!70, 
  fontupper=\footnotesize, 
  coltitle=white, 
  coltext=black, 
  boxrule=0.5mm, 
  title=GPT Assisted Evaluation Prompt
]

\#\# Role\\
You are an impartial evaluator. Your task is to assess whether a model-generated response accurately and coherently matches a human-annotated reference answer.\\

Each input contains two structured components:\\
- <think>: the reasoning or analytical explanation\\
- <conclusion>: the final judgment (e.g., real or fake)\\

\#\# Evaluation Dimensions\\
You should compare the \textbf{ModelOutput} to the \textbf{GroundTruth}, and assign integer scores from 1 to 5 (no decimals) for the following four dimensions:\\

1. Completeness\\
- Does the ModelOutput address all aspects covered in the GroundTruth?\\
- More complete responses should include all relevant information, especially key \"golden clues\".\\
- Incomplete or partially aligned answers should receive lower scores.\\

2. Relevance\\
- Does the ModelOutput discuss the same detection dimensions as in the GroundTruth?\\
- Temporal features include:\\
\hspace*{1em}- Luminance discrepancy\\
\hspace*{1em}- Duplicated components\\
\hspace*{1em}- Awkward facial expressions\\
\hspace*{1em}- Motion inconsistency\\
- Spatial features include:\\
\hspace*{1em}- Abnormal texture\\
\hspace*{1em}- Distorted or omitted components\\
\hspace*{1em}- Chromatic irregularity\\
\hspace*{1em}- Impractical luminosity\\
\hspace*{1em}- Localized blur, etc.\\
- Penalize if irrelevant aspects are introduced or relevant ones are missing.\\

3. Level of Detail\\
- Does the ModelOutput describe fine-grained visual cues in each dimension?\\
- High scores require specific subcomponent elaboration, not just general terms.\\
- Penalize vague or generic responses that lack specific observations.\\

4. Explanation\\
- Is the reasoning in \verb|<think>| logically consistent with the \verb|<conclusion>|?\\
- The explanation should provide clear, causally-linked justifications.\\
- Penalize if the conclusion contradicts the reasoning or lacks support.

\end{tcolorbox}
\caption{GPT Assisted Evaluation Prompt}
\label{prompt:gpt_eva}
\end{figure*}

\section{Training Detail}
\label{apd:training}

The RL training process of \baseline~is illustrated in Fig.\ref{fig:training}. Our GRPO training exhibits a stable increase in reward, eventually reaching 0.95. However, we also observe that both the clip ratio and the maximum response length surge sharply around 200–300 steps, suggesting that a portion of the training samples may be overly challenging for the model.

\section{Case Study: Qualitative Comparison of Methods}
\label{app:case}

\begin{figure*}[ht]
    \centering
    \includegraphics[width=0.95\textwidth]{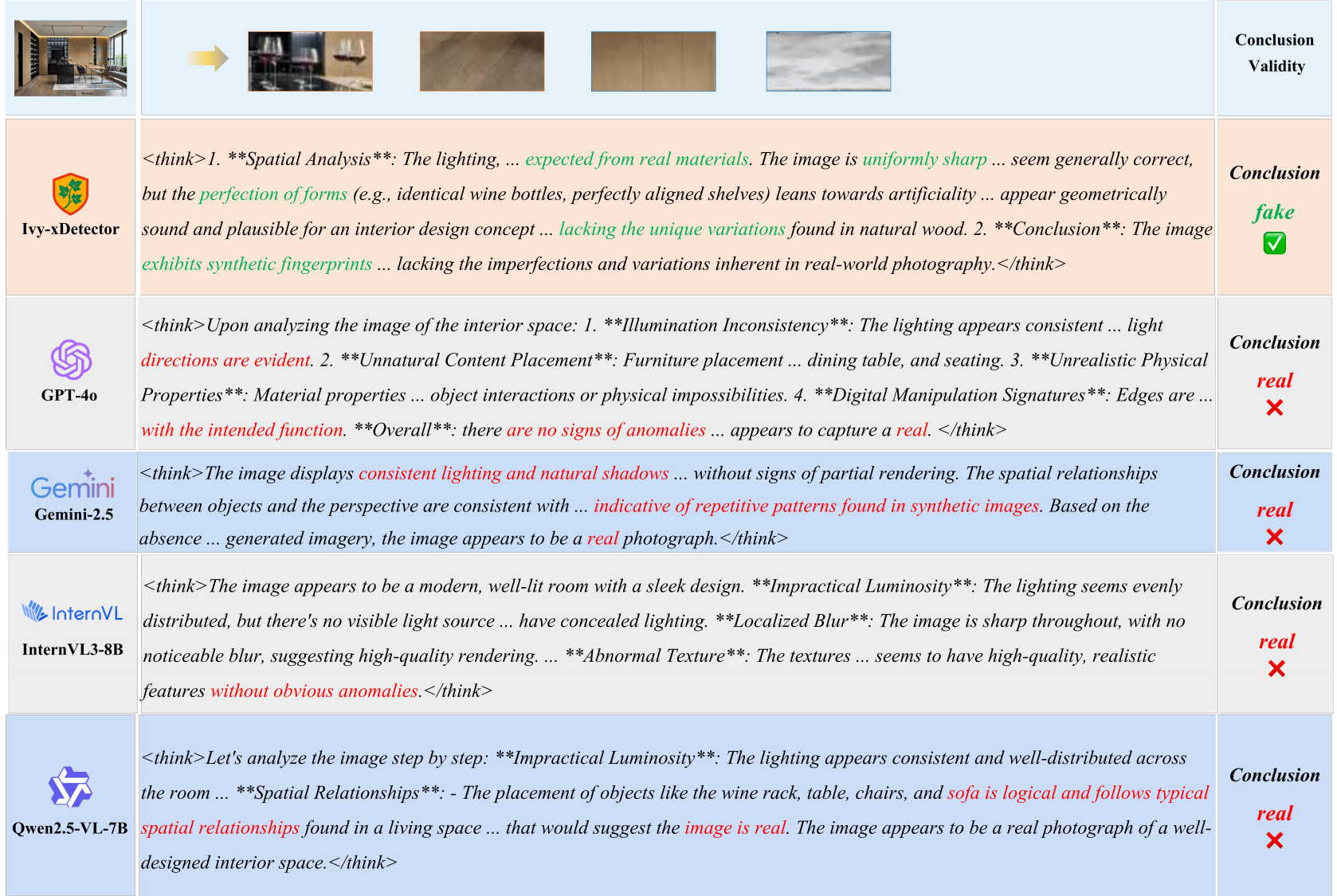}
    \caption{\textbf{Image} example 1 where \textbf{Ivy-xDetector} successfully detects subtle spatial anomalies missed by baselines.}
    \label{fig:case1-image}
\end{figure*}

\begin{figure*}[ht]
    \centering
    \includegraphics[width=0.95\textwidth]{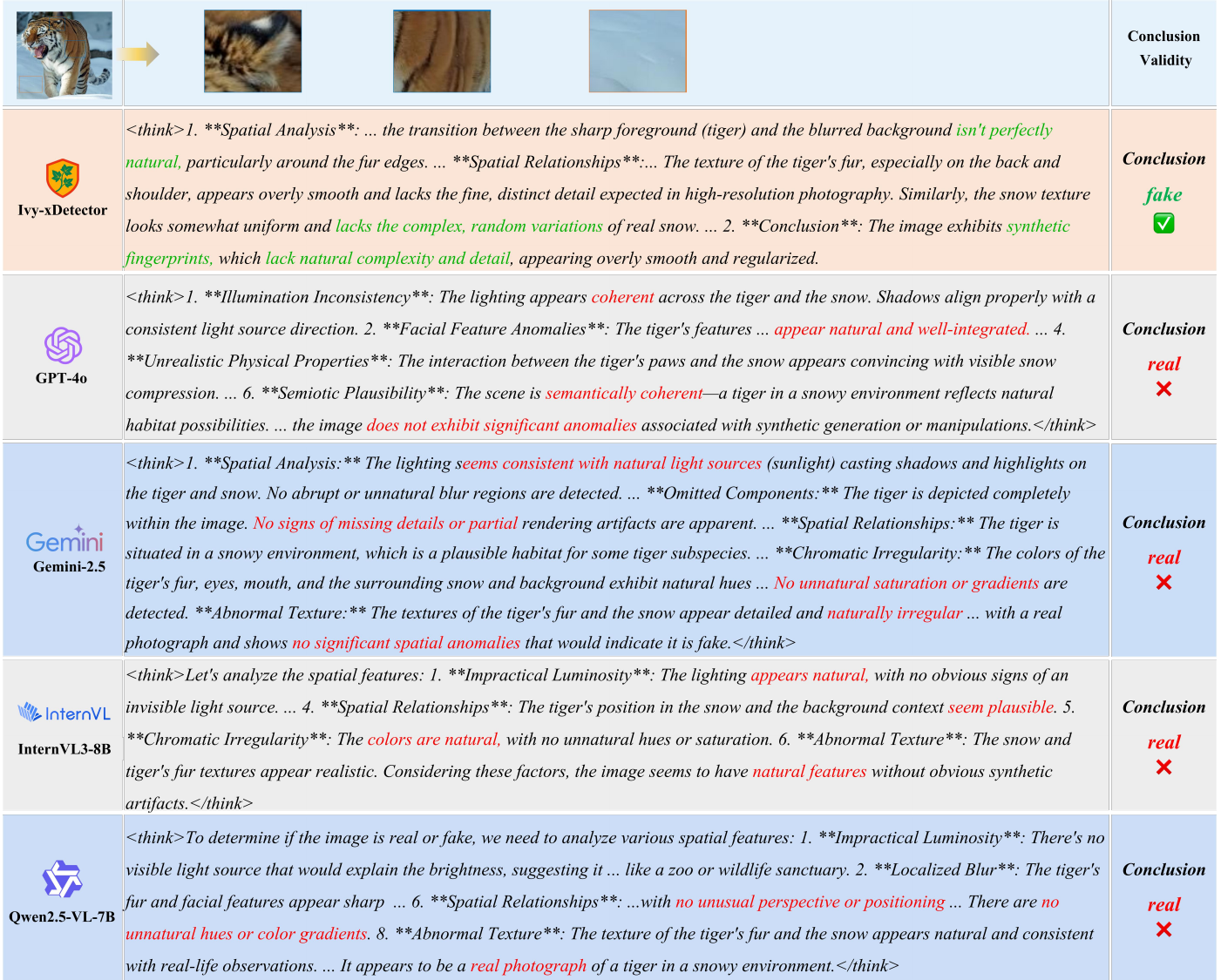}
    \caption{\textbf{Image} example 2 illustrating improved robustness of \textbf{Ivy-xDetector} against visually deceptive artifacts.}
    \label{fig:case2-image}
\end{figure*}

\begin{figure*}[ht]
    \centering
    \includegraphics[width=0.95\textwidth]{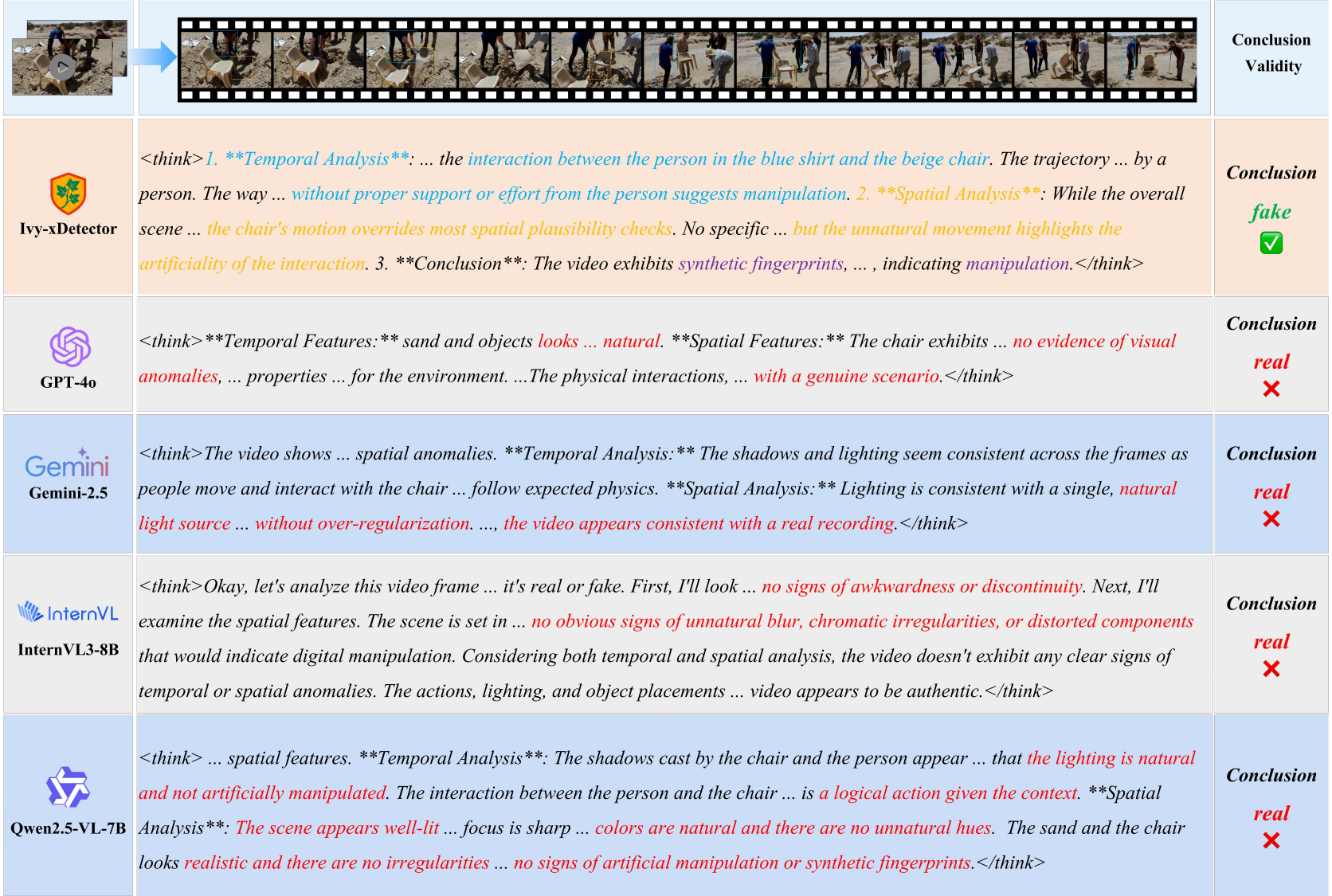}
    \caption{\textbf{Video} example 1 showcasing \textbf{Ivy-xDetector}'s superior ability to detect subtle cross-frame temporal artifacts.}
    \label{fig:case3-video}
\end{figure*}

\begin{figure*}[ht]
    \centering
    \includegraphics[width=0.95\textwidth]{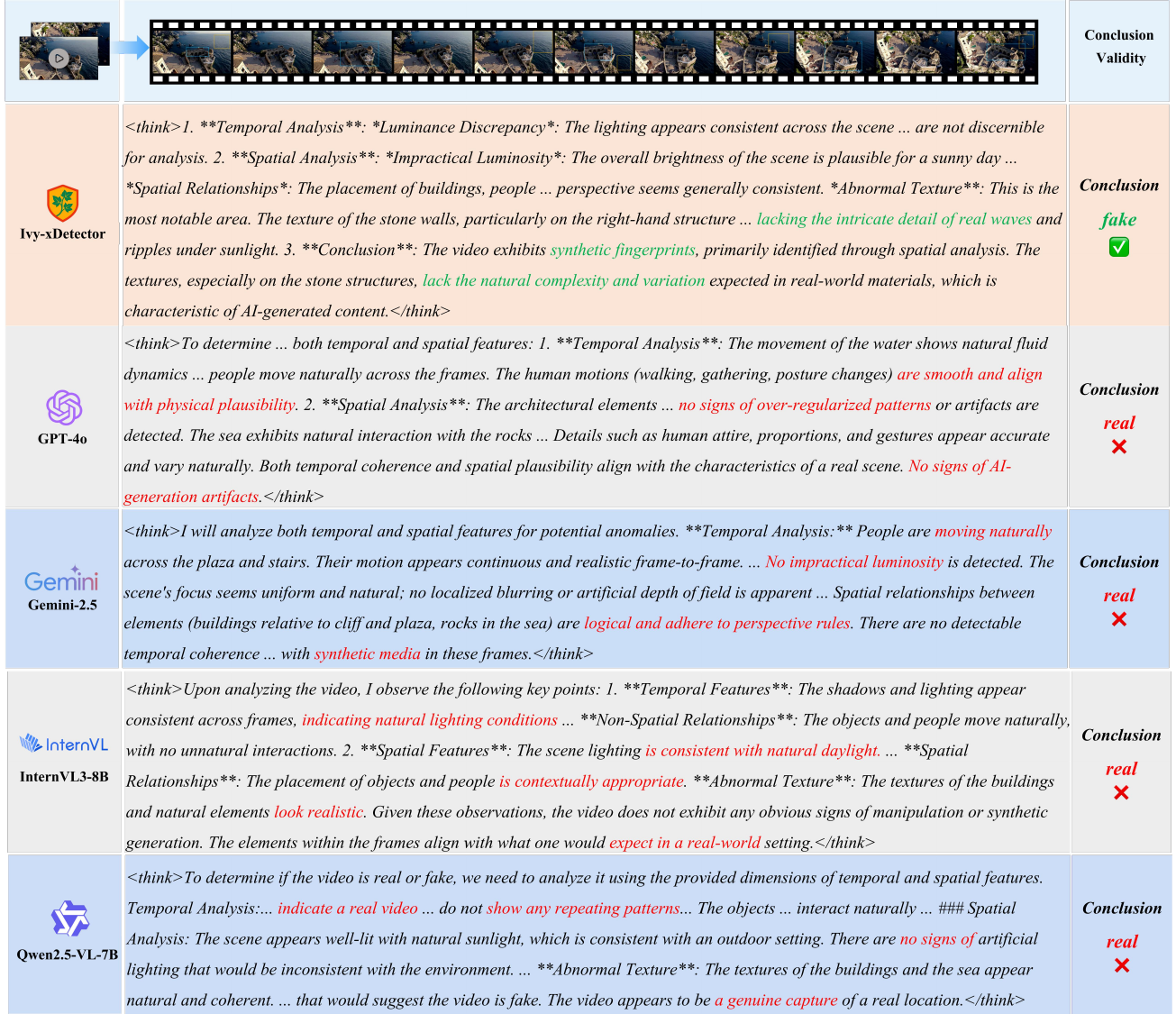}
    \caption{\textbf{Video} example 2 showcasing \textbf{Ivy-xDetector}'s superior ability to detect subtle cross-frame temporal artifacts.}
    \label{fig:case4-video}
\end{figure*}

According to Figures~\ref{fig:case1-image},~\ref{fig:case2-image},~\ref{fig:case3-video}, and~\ref{fig:case4-video}, \baseline~consistently demonstrates superior performance in detecting both spatial and temporal anomalies, providing stronger generalization and robustness compared to existing baselines.

\end{document}